\documentclass[10pt,journal,compsoc]{IEEEtran}

\usepackage{fancyhdr}
\usepackage[normalem]{ulem}
\usepackage{graphicx}
\usepackage{colortbl}
\usepackage{enumitem}
\usepackage{lipsum}
\usepackage{amsmath}
\usepackage{flushend}
\usepackage{subfig}
\usepackage{tikz}
\usepackage{color}
\usepackage{soul}
\usepackage{setspace}
\usepackage{multirow}
\usepackage{tablefootnote}
\usepackage{threeparttable}
\usepackage{xspace}
\usepackage{listings}
\usepackage[noend]{algpseudocode}
\usepackage{mathtools}
\usepackage{gensymb}
\usepackage{booktabs}
\usepackage{hyperref}

\usepackage[short,nodayofweek,level,12hr]{datetime}

\usepackage[noadjust]{cite}

\DeclareRobustCommand\circled[1]{\tikz[baseline=(char.base)]{
           \node[shape=circle,draw,inner sep=0pt,fill=black, text=white] (char) {#1};}}
\DeclarePairedDelimiter\floor{\lfloor}{\rfloor}
\definecolor{dkgreen}{rgb}{0,0.6,0}
\definecolor{gray}{rgb}{0.5,0.5,0.5}
\definecolor{mauve}{rgb}{0.58,0,0.82}
\definecolor{Moegi}{rgb}{0.357, 0.537, 0.188}

\fancypagestyle{arXivstyle}
{
   \fancyhf{}
   \fancyfoot[C]{\vspace{0.02cm}\normalsize \thepage}
}

\makeatletter
\def\BState{\State\hskip-\ALG@thistlm}
\makeatother

\newif\ifsubmission
\submissiontrue

\ifsubmission
\newcommand\lois[1]{#1}

\newcommand\todo[1]{}

\else

\newcommand\lois[1]{{\color{blue}#1}}%
\newcommand\todo[1]{{\color{red} {\bf \fbox{TODO: }}{\it#1}}}

\fi

\newcommand{\mechanism}[1]{EcoFlow}

\AtBeginDocument{%
  \providecommand\BibTeX{{%
    \normalfont B\kern-0.5em{\scshape i\kern-0.25em b}\kern-0.8em\TeX}}}

\newif\ifcameraready
\camerareadytrue

\submissiontrue

\begin{document}
\bstctlcite{IEEEexample:BSTcontrol} %

\definecolor{MidnightBlue}{rgb}{0.1, 0.1, 0.44}
\newcommand{\versionnum}[0]{1.4~---~\today~@~\currenttime~CET} 
\fancyhead{}
\ifcameraready
 \thispagestyle{arXivstyle}
 \pagestyle{arXivstyle}
\else
 \fancyhead[C]{\textcolor{MidnightBlue}{\emph{arXiv - 2022 submission \versionnum{}}}}%
 \fancypagestyle{firststyle}
 {
  \fancyhead[C]{\textcolor{MidnightBlue}{\emph{arXiv - 2022 submission \versionnum{}}}}%
  \fancyfoot[C]{\thepage}
 }
 \thispagestyle{firststyle}
 \pagestyle{firststyle}
\fi

\title{\mechanism{}: Efficient Convolutional Dataflows\\for Low-Power Neural Network Accelerators}
\date{}

\author{Lois~Orosa, Skanda~Koppula, Yaman~Umuroglu, Konstantinos~Kanellopoulos,\\ Juan\ G\'omez-Luna, Michaela~Blott, Kees~Vissers, Onur~Mutlu
\IEEEcompsocitemizethanks{\IEEEcompsocthanksitem Lois Orosa, Konstantinos Kanellopoulos, Juan\ G\'omez-Luna and Onur~Mutlu are with ETH Zurich.%
\IEEEcompsocthanksitem Skanda Koppula is with DeepMind.%
\IEEEcompsocthanksitem Yaman~Umuroglu,  Michaela~Blott, and Kees~Vissers are with Xilinx.}%
}

\IEEEtitleabstractindextext{%

\begin{abstract}
Dilated and transposed convolutions are widely used in modern convolutional neural networks (CNNs). These kernels are used extensively during CNN training and inference of applications such as image segmentation and high-resolution image generation. Although these kernels have grown in popularity, they stress current compute systems due to their high memory intensity, exascale compute demands, and large energy consumption.

We find that commonly-used low-power CNN inference accelerators based on spatial architectures are \emph{not} optimized for both of these convolutional kernels. Dilated and transposed convolutions introduce significant zero padding when mapped to the underlying spatial architecture, significantly degrading performance and energy efficiency. Existing approaches that address this issue require significant design changes to the otherwise simple, efficient, and well-adopted architectures used to compute direct convolutions.

To address this challenge, we propose \mechanism{}, a new set of dataflows and mapping algorithms for dilated and transposed convolutions. These algorithms are tailored to execute efficiently on existing low-cost, small-scale spatial architectures and requires minimal changes to the network-on-chip of existing accelerators. At its core, \mechanism{} eliminates zero padding through careful dataflow orchestration and data mapping tailored to the spatial architecture. \mechanism{} enables flexible and high-performance transpose and dilated convolutions on architectures that are otherwise optimized for CNN inference.

We evaluate the efficiency of our dataflows on CNN training workloads and Generative Adversarial Network (GAN) training workloads. Experiments in our new cycle-accurate spatial architecture simulator show that \mechanism{} 1) reduces end-to-end CNN training time between 7-85\%, and 2) improves end-to-end GAN training performance between 29-42\%, compared to state-of-the-art CNN inference accelerators.

{\bf  [Open-Source Artifact]} We \emph{open-source} both our Spatial Architecture Simulator for Machine Learning (SASiML) and the SASiML compiler to help enable the development of new dataflows and high-accuracy simulation environments for new spatial architectures and dataflows. This can be freely found at \url{https://github.com/CMU-SAFARI/sasiml}.
\end{abstract}

}

\maketitle
\thispagestyle{arXivstyle}

\section{Introduction}

Deep convolutional neural networks (CNNs) have been widely adopted to solve hard problems in computer vision, natural language understanding, speech processing, medical applications, and more~\cite{hinton2012deep,resnet,kalchbrenner2014convolutional,deeplearning,medicine, yolo,goodfellow2014generative,medicalgan,pix2pix,wavegan,cyclegan,videogan,khan2020coronet,shen2015multi,cheng2016computer,kim2016deep,li2014deep,feng2019using,tian2018deeptest,lindsey2018deep,hannun2019cardiologist,sharma2019hiding,hermann2020deep,tian2018deep,schutt2019unifying,lu2019deepvcp,gao2019bidirectional,li2018deepnis,su2020deep,lubbers2018hierarchical,zhu2019phasenet,kaya2019analysis,jain2019extended,zhang2018detecting,zhang2019deep,julian2019deep,shafique2020robust}. Transposed and dilated convolutions are the two key workhorses used to train CNNs, and run a variety of other deep learning models~\cite{bengio2012}. For example, both kernels are employed in applications requiring significant upsampling or downsampling to process high-resolution media such as image generation (using Generative Adversarial Networks (GANs) and Variational Auto-encoders (VAEs)~\cite{goodfellow2014generative, kingma2013auto}), image super-resolution~\cite{superresolution,shi2017single,zhang2018dcsr}, and image segmentation~\cite{girshick2015fast,ren2015faster}. Additionally, more emerging machine learning works in text-to-speech generation~\cite{van2016heiga}, speech recognition~\cite{han2019state}, and audio synthesis~\cite{pons2020upsampling} use dilated convolutions. Other experimental machine learning models, such as hierarchical capsule networks~\cite{srivastava2019hierarchical} and dilated residual networks~\cite{yu2017dilated} for improved image modeling, use both these convolution types.

Meanwhile, specialized architectures for CNN inference have gained traction to support the demand for low-cost deep learning on a variety of devices~\cite{Chen2017,Chen16}: Internet-of-Things devices (IoT)~\cite{xu2018scaling,liu2019edge,lane2018deep,hadidi2019characterizing}, phones, wearables~\cite{mathur2017deepeye}, servers, and various embedded electronics~\cite{lane2017squeezing,andri2017yodann}.
While these works demonstrate efficient execution of direct convolutions (i.e., regular or `standard' convolutions), we find that existing dataflows for transposed and dilated convolutions are poorly tailored for these architectures, causing significant bottlenecks for emerging edge workloads that use transpose and dilated convolutions. Despite this issue, these workloads are of growing interest to manufacturers, because they can enable: (1) on-device model training for improved user data privacy~\cite{ondevicelearning,stubbs2017physical,teerapittayanon2017distributed}, (2) high-resolution image generation critical for augmented reality~\cite{ranjan2015natural,donahue2019large}, (3) real-time speech recognition and generation~\cite{van2016heiga,zhang2018dcsr}, and many other applications employing dilated and transposed convolutions~\cite{long2015fully,chen2017rethinking,yu2017dilated,han2019state, pons2020upsampling,superresolution,shi2017single,arik2018fast,li2019inversion,linmans2018sample,gu2021facial,yang2021transpose,zhang2021scn,lim2018foreground,zhou2018d,wang2019deeply,liu2019dilated,deb2018aggregated,chang2018temporal,lim2018foreground}.
 
To address this issue, we introduce \textit{\mechanism{}}, a new set of dataflows and data mappings designed to efficiently perform transposed and dilated convolutions on low-cost, small-scale spatial architectures that are already widely in-use for regular CNN inference. We identify key bottlenecks introduced by these operations, originating from padding and zero-insertions required to up- and down-sample feature maps. \mechanism{} circumvents these bottlenecks by meticulously orchestrating the data mapping and dataflows onto the target spatial architecture. By eliminating unnecessary operations, \mechanism{} achieves significant improvements in performance and energy consumption, with minimal changes to the spatial array of a common CNN inference accelerator.

We improve on several prior works that propose specialized accelerators that target specifically either transposed convolutions~\cite{zara,mao2018lergan,im2020dt,im2019dt}, dilated convolutions \cite{im2020dt,im2019dt}, or general sparsity~\cite{song2018towards, zhu2020efficient, cambriconx, kanellopoulos2019smash, mishra2017fine, pal2018outerspace, umuroglu2014energy, cui2016towards}. We generalize, simplify, and significantly reduce the required architectural changes to support exactly the structured sparsity of these convolutional kernels. Our design goal is to avoid highly-specialized accelerator architectures that are markedly different from common and well-understood spatial architectures (i.e., a matrix of processing elements working in a systolic array fashion) optimized for direct convolutions. Re-use of existing hardware architectural designs permits lower testing and manufacturing costs. \lois{\mechanism{} could also inspire the optimization of dataflows for spatial architectures designed to accelerate other applications such as genome sequence analysis~\cite{cali2020genasm,alser2020accelerating,senol2019nanopore,singh2021fpga,xin2013accelerating,alser2017gatekeeper,kim2018grim}.}

We make the following key contributions:
\begin{itemize}
\itemsep0em

\item We propose \mechanism{}, a new set of dataflows and data mappings that enable efficient execution of transpose and dilated convolutions on CNN inference accelerators by introducing minimal hardware changes (Section~\ref{sec:idea}). 

\item We develop a cycle-accurate spatial architecture simulator to evaluate \mechanism{}. Our architectural simulator includes TPU~\cite{jouppi2017datacenter}, Eyeriss~\cite{Chen2017}, and \mechanism{} models, and it supports efficient execution of transposed, dilated, and direct convolutions (Section~\ref{sec:sasim}).%

\item We comprehensively evaluate the performance and energy efficiency of \mechanism{}. Our evaluation shows that \mechanism{}: 1) reduces end-to-end CNN training time between 7-85\%, and 2) improves end-to-end GAN training performance between 29-42\%, compared to state-of-the-art CNN inference accelerators.
\end{itemize}

\section{Background}
\label{sec:background}

\label{sec:background_cnn}
A deep convolutional neural network (CNN) is a neural network with one or more convolutional layers. A convolutional layer in a CNN applies a sliding filter to a 2D or 3D matrix that represents the input image or intermediate layer input. The input and output matrices to a convolutional layer are referred to as the \emph{input feature map (ifmap)} and \emph{output feature map (ofmap)}, respectively. The \emph{filter} (or \emph{kernel}) is the sliding filter that is applied to the ifmap to calculate the ofmap. In each convolutional layer, there are usually multiple filters applied in parallel to the ifmap, producing multiple output matrices that compose the ofmap.  
The \text{stride} of a convolution refers to the size of the step that the convolutional filter takes while sliding through the ifmap. 
The core operation of a CNN is a multiply-and-accumulate (MAC) operation. Modern CNNs may have up to $10^{18}$ MACs performed during one forward evaluation \cite{imagenetminutes}. Most of these MAC operations are performed in the convolutional layers~\cite{dnntrain}. 

Inference is the production phase where the CNN classifies unknown images. Before performing inference in production, the network is trained with and algorithm called backpropagation, that includes the calculation of input gradients and filter gradients. For a detailed treatment of CNN operation and gradient computation, we refer the reader to \cite{lecun1990handwritten, krizhevsky2012, vgg, deeplearning, szegedy2015going, russakovsky2015imagenet, schmidhuber2015deep, zeiler2014visualizing, sutskever2014sequence, goodfellow2016deep,bishop, cnnnotes}.

\subsection{Different Types of Convolutions in CNNs}
\label{sec:background_training}

\begin{figure}[h] 
\centering
    \includegraphics[width=0.99\linewidth]{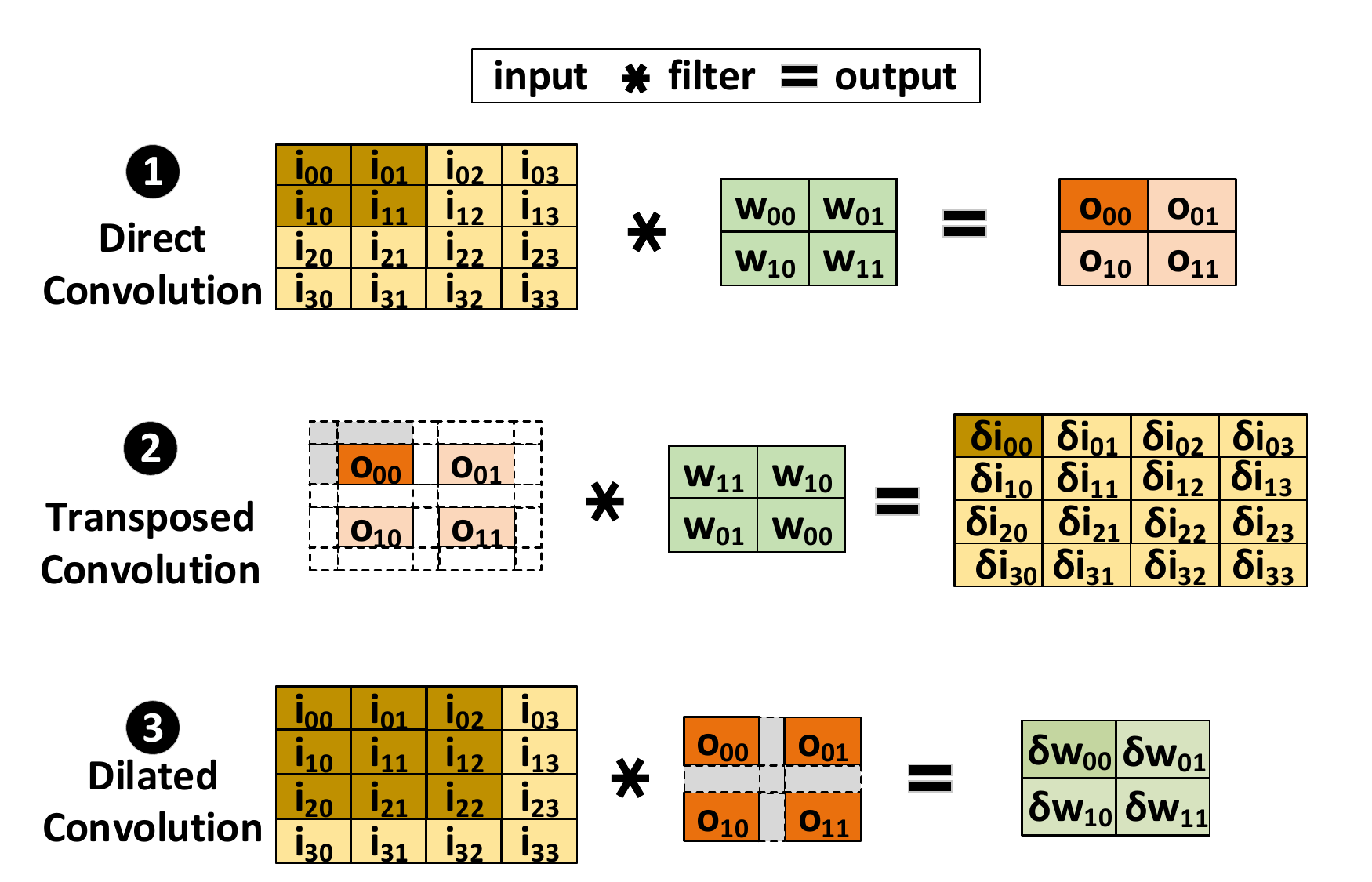}
    \caption{Different types of convolutions with an example 4x4 input, 2x2 filter, and stride 2 used in the CNN training phase.}
    \label{fig:training_steps}
\end{figure}

Figure~\ref{fig:training_steps} illustrates the three main types of convolutions we can find in convolutional neural networks. This examples shows the case for the CNN training phase of a convolutional neural network, where we find direct convolutions in the forward pass, and transposed and dilated convolutions in the backward pass.

\subsubsection{Direct Convolution}
\label{sec:forward}

A direct convolution (also known as convolution, standard convolution, or regular convolution) is one of the most common operations in convolutional neural networks (in both inference and training), and other variants of CNNs~\cite{goodfellow2014generative,van2016heiga,han2019state,pons2020upsampling,srivastava2019hierarchical,yu2017dilated,kingma2013auto,long2015fully,chen2017rethinking,cho2014learning,mao2014deep}. A direct convolution is performed by sliding the filter ($W_{xy}$) over the input ($i_{xy}$) with a specific stride (stride 2 in the example \circled{1} in Figure~\ref{fig:training_steps}), generally starting at the top left corner, so as to move the filter to the boundary of the input (\circled{1} in Figure~\ref{fig:training_steps}).

\subsubsection{Transposed Convolution}
\label{sec:backward}

A transposed convolution operation forms the same connectivity as a direct convolution but in the backward direction, which requires upsampling the input into an output of larger dimensions. Transposed convolutions are commonly used in CNN training and in emerging CNN  workloads~\cite{long2015fully,chen2017rethinking,yu2017dilated,han2019state, pons2020upsampling,superresolution,shi2017single,   yang2021transpose,zhang2021scn,lim2018foreground,gu2021facial,li2019inversion,linmans2018sample}. 
Figure~\ref{fig:training_steps} \circled{2} shows an example that calculates the input gradients ($\delta i_{xy}$) in the backward propagation pass of CNN training. A transposed convolution is computed by convolving the error matrix ($O_{xy}$) with the forward pass filter ($W_{xy}$) rotated 180\degree. Transposed convolutions introduce zero padding into the error matrix to produce an output of larger dimensions, up-sampling the backpropagated errors. %
The error matrix might require zero-padding in the borders, as in Figure~\ref{fig:training_steps} \circled{2}. If the stride is greater than one (the example Figure~\ref{fig:training_steps} \circled{2} has stride 2), the error matrix also require internal zero padding as zero-valued rows and columns. 

\subsubsection{Dilated Convolution}
\label{sec:backward}
 Dilated convolutions are commonly used in CNN training and in emerging CNN  workloads~\cite{long2015fully,chen2017rethinking,yu2017dilated,han2019state, pons2020upsampling,superresolution,shi2017single, liu2019dilated,deb2018aggregated,chang2018temporal,wang2019deeply,zhou2018d}. Figure~\ref{fig:training_steps} \circled{3} shows a dilated convolution example that calculates the filter gradients ($\delta W_{xy}$) with dilation rate = 2 (i.e., stride 2) in the backward propagation pass of CNN training. A dilated convolution is computed by convolving the input ($i_{xy}$) with a padded filter ($O_{xy}$) to augment its dimensions. This convolution inserts zero padding as rows and columns in the filter when the dilatation rate (i.e., the stride of the convolution when training a CNN) is greater than one. A dilation rate of 1 does not introduce any padding in the filter.

\subsection{Spatial Architectures for CNN Inference}
\label{sec:background_spatial_architectures}

A spatial compute array is the key component in many popular low-cost CNN accelerators \cite{Chen2017, eyerissv2, flexflow, flexflow2, flexflow3, jouppi2017datacenter,jouppi2021ten,khabbazan2019design,andri2017yodann,wei2018tgpa,choi2018low,wang2019flexible,hu2019resources,venkataramani2021rapid}. A spatial architecture consists of a matrix of simple processing elements (PEs), interconnected with one or several internal networks. Each PE is able to perform a MAC operation. By orchestrating data into and out of the PE network, spatial architectures can efficiently implement either matrix multiplications or convolutions. Examples of spatial architectures include Eyeriss V1/V2~\cite{Chen2017, eyerissv2}, Google's TPU~\cite{jouppi2017datacenter,jouppi2021ten}, NVIDIA's CUDA Tensor Cores~\cite{markidis2018nvidia}, Nanofabrics~\cite{goldstein2001nanofabrics}, TRIPS~\cite{Sankaralingam2003}, RAW~\cite{lee1998space}, SmartMemories~\cite{mai2000smart}, FlexFlow~\cite{flexflow, flexflow2, flexflow3}, SCNN~\cite{scnn}, and Morph~\cite{morph}.

\begin{figure}[h] 
\centering
    \includegraphics[width=0.85\linewidth]{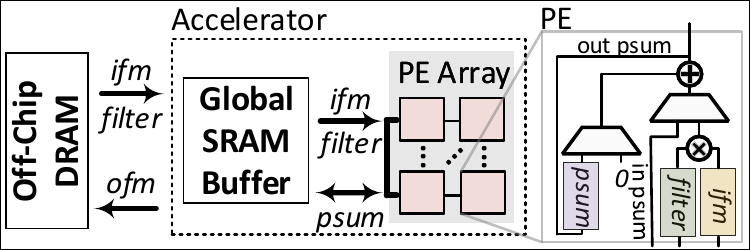}%
    \caption{Common Spatial architecture for CNN inference acceleration.}
    \label{fig:eyerissSec2}
\end{figure}

Figure~\ref{fig:eyerissSec2} illustrates the core elements of a common spatial architecture for CNN inference. At the core is an array of interconnected PEs. Data is cached on a global on-chip buffer, which utilizes various network-on-chips (NoCs) to exchange data with the PE array. In common designs \cite{Chen2017,flexflow}, this network enables data transfer between vertically adjacent PEs, simultaneous broadcast to all PEs, and multicasting values to individual sets of PEs. On the left of Figure~\ref{fig:eyerissSec2} we can see the off-chip memory that stores temporary data that overflows the global buffer, and the complete set of ifmaps, filters, and final ofmaps. 
The internal architecture of the PEs (right side of Figure~\ref{fig:eyerissSec2}) can differ slightly, based on the chosen dataflow, accelerator function (e.g. sparse/non-sparse CNN acceleration), and other optimizations (e.g. reduced precision, clock-gating). 
PEs generally store small amounts of weight or partial sum data which is reused during dataflow~\cite{Chen2017, tetris}.

\subsection{CNN Dataflows on Spatial Architectures}
\label{sec:dataflow_conv}
We describe the most widely-used dataflows for performing convolutions in spatial architectures, used to evaluate \mechanism{} in Section~\ref{sec:evaluation}. An in-depth discussion of each dataflow can be found in \cite{Chen2017}.

\vspace{5pt}\noindent\textbf{Convolution Dataflows.} 
Row stationary (RS)~\cite{Chen2017,zhang2021novel} is a state-of-the-art dataflow for performing convolutions in spatial architectures. The RS dataflow attempts to minimize the overall energy consumed by off-chip data accesses by re-using the convolutional filters and ifmaps. RS minimizes data movement across all data types by effectively assigning each PE a 1D convolution to perform. The results of these 1D convolutions (or partial sums) are accumulated with other partial sums from other PEs to produce the final ofmap. RS %
has been shown to be the most energy efficient dataflow on spatial architectures~\cite{Chen2017}, compared to Weight Stationary (WS)\cite{sankaradas2009massively, yoo20151, cavigelli2015origami} and Output Stationary (OS)~\cite{du2015shidiannao, peemen2013memory, gupta2015deep} dataflows.

Although previous works claim that the choice of dataflow is not critical for direct convolutions~\cite{yang2018dnn}, in this work we demonstrate that this choice \emph{does} matter for transposed and dilated convolutions. Using direct convolution dataflows for transposed and dilated convolutions can result into low performance and poor energy efficiency.

\vspace{5pt}\noindent\textbf{Matrix Multiplication Dataflows.}
Lowering a convolution into a matrix multiplication is a well known technique that is used today in many CNN frameworks and accelerators, i.e., TPUs~\cite{jouppi2017datacenter,jouppi2021ten}. For a detailed explanation of the lowering process, we refer the reader to~\cite{chetlur2014cudnn}. After lowering, several dataflows can be used for the matrix multiplication~\cite{kung1979systolic}. A common approach uses an output stationary dataflow in which partial sums are accumulated locally, and inputs are forwarded to adjacent rows \cite{Chen2017}. The matrices are fed into the PE array from the top and left edges of the array \cite{eyerissv2}. This is the approach used in our reference implementation in Section~\ref{sec:evaluation}.

\section{Motivation and Goal}
\label{sec:challenges}

We describe the main inefficiencies of transpose and dilated convolutions, and how related works require a specialized accelerators to solve this problem (Section~\ref{sec:inefficiency_grad_calc}). Our goal in this paper is to introduce minimal changes to an existing DNN inference accelerator to perform tranpose and dilated convolutions (Section~\ref{sec:goal}).

\subsection{Inefficiencies of Transposed and Dilated Convolutions}
\label{sec:inefficiency_grad_calc}

\begin{figure}[h] 
    \centering
    \includegraphics[width=0.75\linewidth]{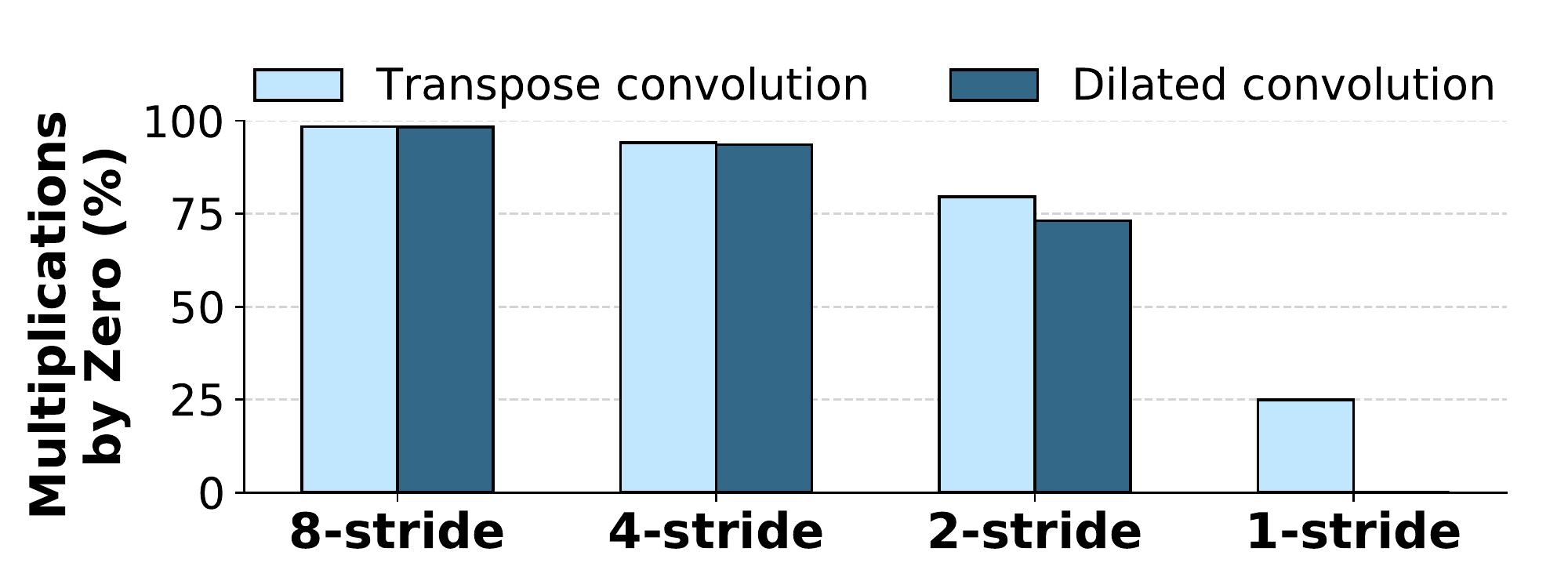}
    \caption{Padding-induced zero multiplications in transpose and dilated convolutions during input and filter gradient calculation of representative CNN layers with different strides.}
    \label{fig:igrad_zeros}
\end{figure}

To understand the mechanics and bottlenecks of transpose and dilated convolution, we analyze the backward pass of CNN training on representative convolutional layers with different strides from two common CNNs, ResNet-50 \cite{resnet} and AlexNet~\cite{krizhevsky2012}.
Figure~\ref{fig:igrad_zeros} shows the percentage of multiplications by zero required to compute both transposed and dilated convolutions. We observe that for strides larger than 1, the zero multiplications dominate utilization by large margins. For example, more than 70\% of multiplications for 2-stride convolutions are zero. The larger the stride, the larger the number of zero multiplications.

We make two observations. First, the PEs that execute zero operations \emph{cannot} be used to perform useful operations, which causes resource under-utilization. Second, although the result of the multiplication is zero, inputs coming from other PEs might need to be accumulated and transmitted to the next node, which practically increases the latency of useful computations and reduce performance.

\subsubsection{Analyzing Transpose Convolutions}
\label{sec:igrad_challenge}

\begin{figure}[h] 
\centering
    \includegraphics[width=0.99\linewidth]{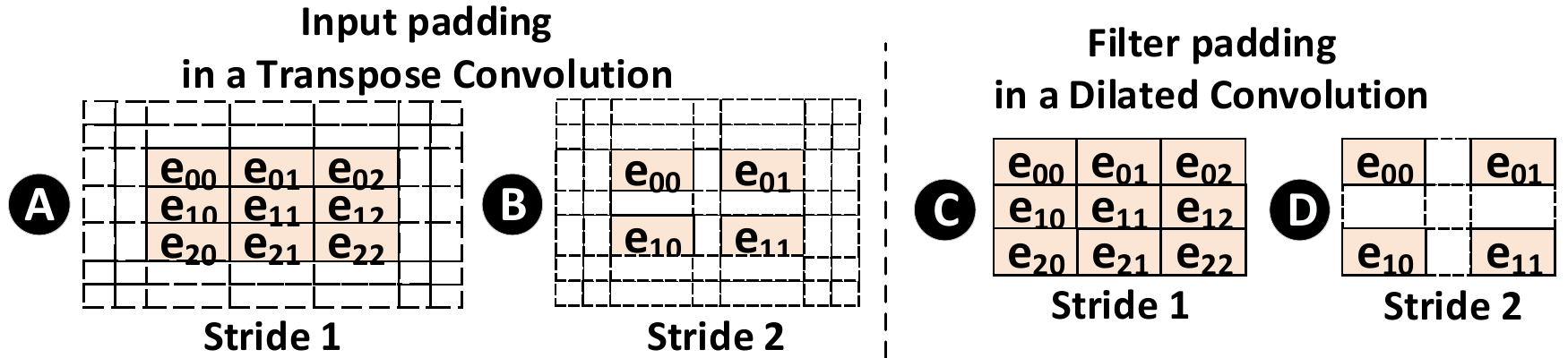}%
    \caption{Example of the zero-padding required to calculate transpose and dilated convolutions.}
    \label{fig:padding-examples-fgrad-igrad}
\end{figure}

Performing a transposed convolution in a spatial architecture designed for CNN inference requires significant padding to obtain the correct ofmap dimensions (i.e., up-sampling).
Figure~\ref{fig:padding-examples-fgrad-igrad} shows two examples of the required padding in the input for obtaining the desired up-sampled ofmap\footnote{The higher the stride, the higher the up-sampling.}. In the example, layer \circled{A} requires 40 outer padding elements in the inputs (81\% of the matrix), and layer \circled{B} requires 40 outer padding elements and 5 inner padding elements in the inputs (92\% of the matrix).

We can formulate the amount of padding required by a particular transposed convolution by considering ifmap, stride, filter sizes. For a $N \times N$ ifmap, $K \times K$ filter, and stride $S$, the number of inner padding elements is given by $\left[S\left(N-1\right) + 1\right]^2-N^2$. The number of outer padding elements is given by $4\left(K-1\right)\left[S\left(N-1\right)+1\right]+4\left(K-1\right)^2$. The total number of zero-padding elements increases \emph{linearly with the ifmap size}, and \emph{quadratically with the stride}.

Transposed convolutions are used for upsampling a input to produce a high-resolution output feature map or media. For example, semantic segmentation \cite{noh2015learning} and super-resolution  \cite{superresolution} CNNs output images that are of the same or higher resolution than their input. Generative Adversarial Networks \cite{dcgan, wavegan, videogan} use transposed convolutions for the same purpose.

\vspace{5pt}\noindent\textbf{{Existing proposals}}. There are several works that propose to accelerate transposed convolutions with specialized GAN accelerators~\cite{song2018towards,im2020dt,im2019dt,ganax,shi2019huge2}. Although these works achieve significant performance and energy improvements, they do it at the cost of designing a specialized accelerator for GANs instead of maintaining a simple, efficient, and more general spatial architecture optimized for CNN inference.

\subsubsection{Analyzing Dilated Convolutions}
\label{sec:fgrad_challenge}

Figure~\ref{fig:padding-examples-fgrad-igrad} illustrates two examples (\circled{C} and \circled{D}) of the required filter zero padding in a dilated convolution. Unlike in transposed convolution, the error matrix is only padded internally. In \circled{C}, the stride is one, so the filter gradients can be calculated without padding. When the stride is larger than one, filter gradient calculation requires inner padding. \circled{D} shows an example of this, with stride 2. 56\% of the padded error matrix is zero. The amount of inner padding follows the same trend as above, increasing linearly with the ifmap size and quadratically with the stride.

Dilated convolutions are used in the forward pass of a handful of emerging, state-of-art classification networks~\cite{yu2017dilated, dilatedconv1, dilatedconv2}. Dilated convolutions are also used for aiding visual interpretation of CNNs \cite{deconvolutional}.

\vspace{5pt}\noindent\textbf{{Existing proposals}}. DT-CNN~\cite{im2019dt} proposes an specialized hardware accelerator to perform both transposed and dilated convolutions using delay cells. Unlike \mechanism{}, DT-CNN is a specialized architecture customized for optimizing image segmentation workloads. %

\subsection{Goal}
\label{sec:goal}

Our proposal builds on two key observations: (1) the padding required to perform transposed and dilated convolutions on spatial architectures has a very negative effect on efficiency, and (2) the padding is strictly determined by the characteristics of the convolution and the dimensions of the feature maps and kernel, and thus the location of zero-values is static and deterministic. Our goal in this work is to exploit these two observations in order to (1) eliminate zero padding to avoid low resource occupation, (2) minimize energy and memory requirements, (3) maximize throughput, and (4) introduce minimal changes to the spatial architecture of common CNN inference accelerators. To this end, we develop \mechanism{}.

\section{\mechanism{}}
\label{sec:idea}

We introduce \mechanism{}, a new set of dataflows and data mapping algorithms for calculating transpose convolutions (Section~\ref{sec:input_gradients}) and dilated convolutions (Section~\ref{sec:filter_gradients}) in spatial architectures of CNN accelerators that are optimized for executing direct convolutions. 

The core idea of \mechanism{} is to meticulously orchestrate dataflow and map computation as to avoid zero padding and occupy PEs with only useful operations. \mechanism{} efficiently reuses the spatial architecture used for direct convolutions to execute both transpose and dilated convolutions efficiently. One of the main characteristics of \mechanism{} is that it can be mapped to existing CNN inference spatial architectures with minimal hardware changes, which allows efficient execution of transposed, dilated, and direct convolutions.
\mechanism{} requires small modifications in the network-on-chip to enable efficient data movements without wasting hardware resources.

The dataflow and mapping onto hardware is computed at \emph{compile time}. \mechanism{}'s mapping is more complex than other state-of-the-art dataflows, but this added complexity is a one-time cost during the initial compilation step. The compiler calculates a Finite State Machine (FSM) that is loaded into the PEs to perform the convolutions at runtime. We explain the details of the hardware architecture in Section~\ref{sec:hw_architecture}.

\subsection{Transpose Convolutions}
\label{sec:input_gradients}

In this section, we explain the steps \mechanism{} takes during compilation time (Section~\ref{sec:gradflow_igrad_compilation}) and runtime (Section~\ref{sec:gradflow_igrad_runtime}) to perform transposed convolutions. %

Without loss of generality, we use an example of the transposed convolution that calculates the input gradients in the CNN training algorithm. In this context, the input of the convolution is the padded error (the amount of padding depends on stride in the forward pass), the filter corresponds to the rotated filter from the forward pass, and the output of the convolution are the calculated input gradients.

\begin{figure}[h] 
\centering
    \includegraphics[width=1.0
    \linewidth]{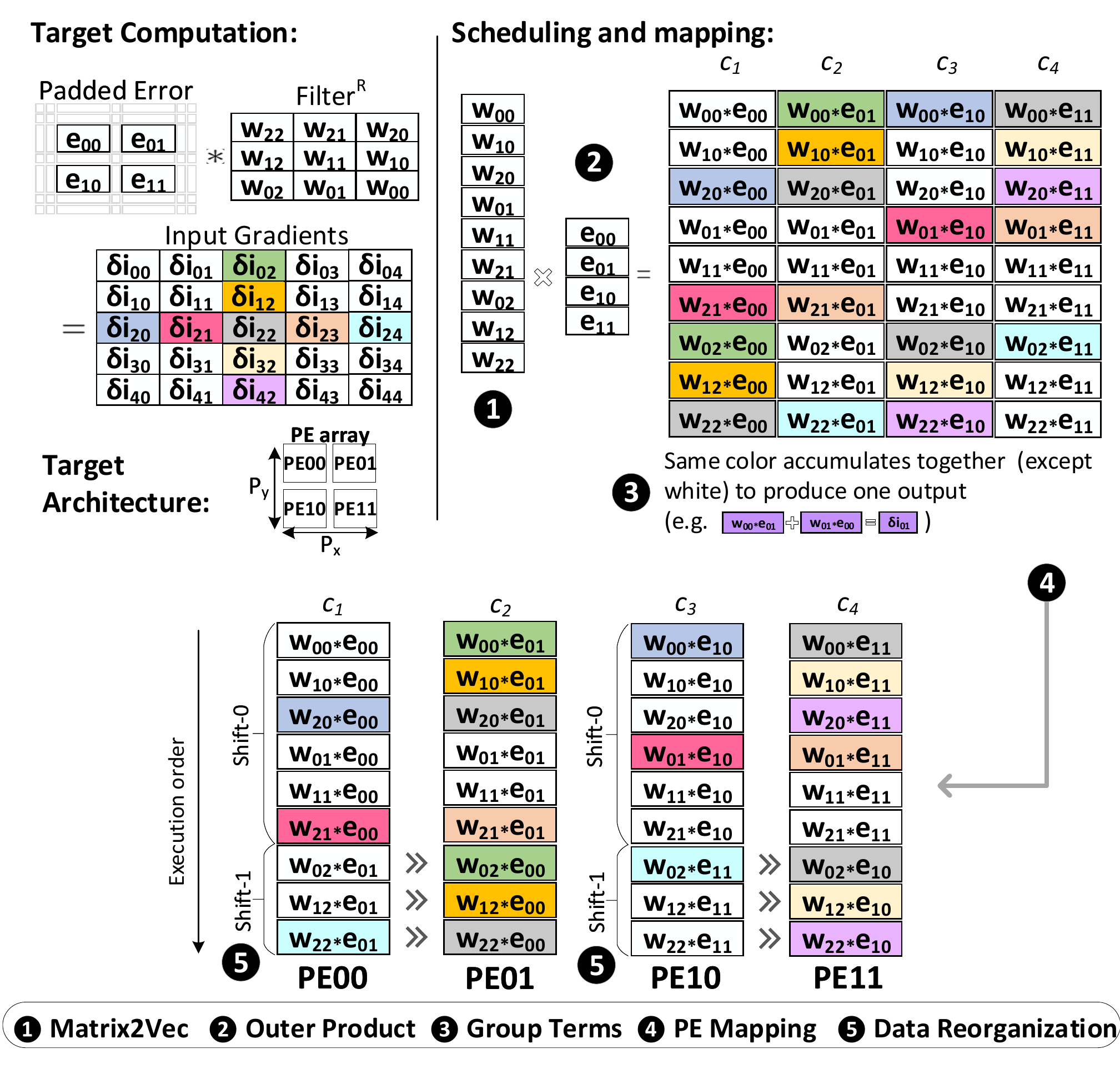}
    \caption{Example of transposed convolution for calculating the input gradients on CNN training algorithm using \mechanism{}. The symbol \textgreater\textgreater~represents a column element \texttt{shift} by one.}
    \label{fig:gradflow-igrad}
\end{figure}

\subsubsection{Compilation Time.}
\label{sec:gradflow_igrad_compilation}
The \mechanism{} compiler determines (1) the computation scheduling required to compute the (transposed) convolution and (2) the mapping of computations onto the architecture's PEs array.

\vspace{2pt}
 \mechanism{} follows five steps to calculate the \textbf{computation scheduling} and \textbf{mapping}. To improve clarity, we walk through each step using the example in Figure~\ref{fig:gradflow-igrad}: a transposed convolution with stride 2, 5$\times$5 output (i.e., input gradients), 3$\times$3 filter (i.e., rotated filter), and 7$\times$7 input (i.e., padded error) reshaped using padding from the original 2$\times$2 error):
 
\vspace{4pt}
\circled{1} The \mechanism{} compiler converts the rotated filter and the error matrix into symbolic vectors. In Figure~\ref{fig:gradflow-igrad}, these vectors have dimensions 9$\times$1 and 4$\times$1, respectively.

\vspace{2pt}
\circled{2} The compiler performs the symbolic outer product of both vectors  by multiplying all elements of the filter by all elements of the error matrix. The resulting matrix contains all multiplications required to perform the transposed convolution for input gradient calculation. Each gradient is the sum of some subset of these products. Notably, this matrix does \emph{not} contain any zero multiplication due to padding. In our example, this matrix has dimension 9$\times$4.

\vspace{2pt}
\circled{3} \mechanism{} determines which matrix elements have to be accumulated together to produce a single input gradient, and marks them with the same \emph{label}.
The labels are determined by doing a transposed convolution with placeholder symbols. In the example, cells with the same color represent matrix elements with the same label. The exception to this are the white cells, which produce a single gradient by themselves; white cells do not need to be accumulated with other values.

\vspace{2pt}
\circled{4} The compiler assigns each column of symbolic computations to a different PE. The mapping assigns consecutive columns to consecutive PEs, from top to bottom and from left to right in the PE array. The number of PEs used by \mechanism{} is equal to the dimensions of the error matrix. In the example, the PE array is composed by 2$\times$2 array, shown in the bottom left. This mapping can be reorganized to reduce the number of required PEs (see \textit{Grouping}). 

\vspace{2pt}
\circled{5} The multiplications are reorganized with the goal of leveraging local point-to-point network to accumulate partial sums across connected, vertically-adjacent PEs. \mechanism{} maps multiplications that must accumulate together either into the same PE, or across vertical PEs. The reorganization consists of \emph{circular shifting} of these multiplication blocks across horizontal PEs. 
Each block shifts $\floor{\frac{w\_{idx}}{W_x \times stride}}$ PEs over, where $W_x$ is one dimension of the filter ($W_x=3$ in the example) and $w\_idx$ is the index of the computation in the order of execution in each PE  (e.g., $w_{00}*e_{00}$ has $w\_idx = 0$, $w_{10}*e_{00}$ has $w\_idx = 1$, etc.). Since the shifting is circular across horizontal PEs, computation blocks in the upper row of PEs shift from PE00 to PE01 and from PE01 to PE00 in the example. In the lower row, computation blocks shift between PE10 and PE11.

In Figure~\ref{fig:gradflow-igrad}, the first six computation blocks are not shifted 
($\floor{\frac{w\_{idx}}{3 \times 2}} = 0$ for $0 \leq w\_{idx} < 6$),
but the next three blocks are shifted over to the horizontally adjacent PE
($\floor{\frac{w\_{idx}}{3 \times 2}} = 1$ for $6 \leq w\_{idx} \leq 9$). 
As a result of this reorganization, all the data that needs to be accumulated together is placed vertically. For example, the light blue multiply operations ($w_{22}*e_{01}$ and $w_{02}*e_{11}$) are shifted so they accumulate across on vertically adjacent PEs, PE00 and PE10.

\begin{figure}[h] 
\centering
    \includegraphics[width=0.99\linewidth]{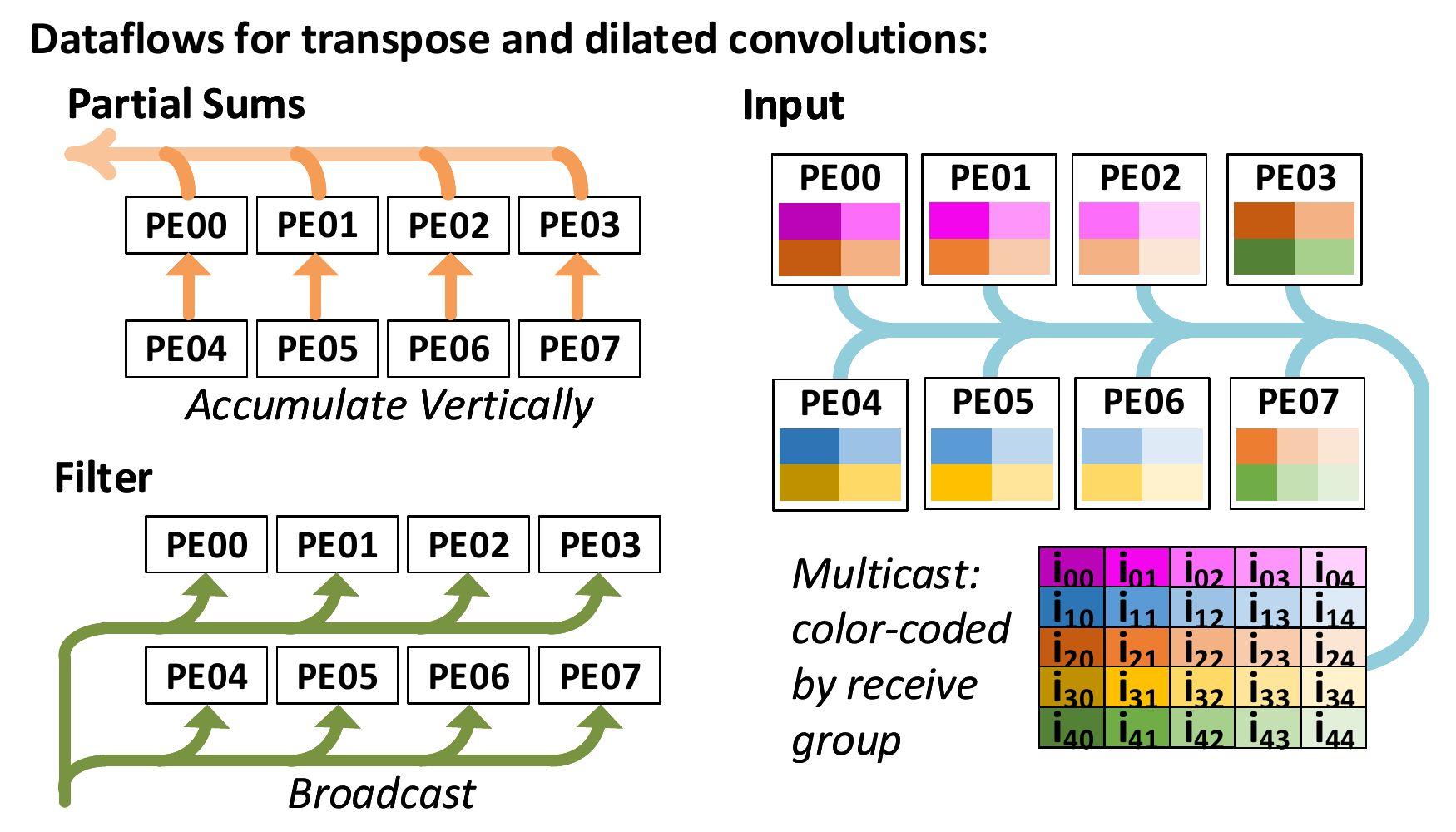}
    \caption{Dataflow for each data type for transpose and dilated convolutions used in CNN training to calculate the input and filter gradients.}
    \label{fig:dataflow}
\end{figure}

\begin{figure}[h] 
\centering
    \includegraphics[width=0.99\linewidth]{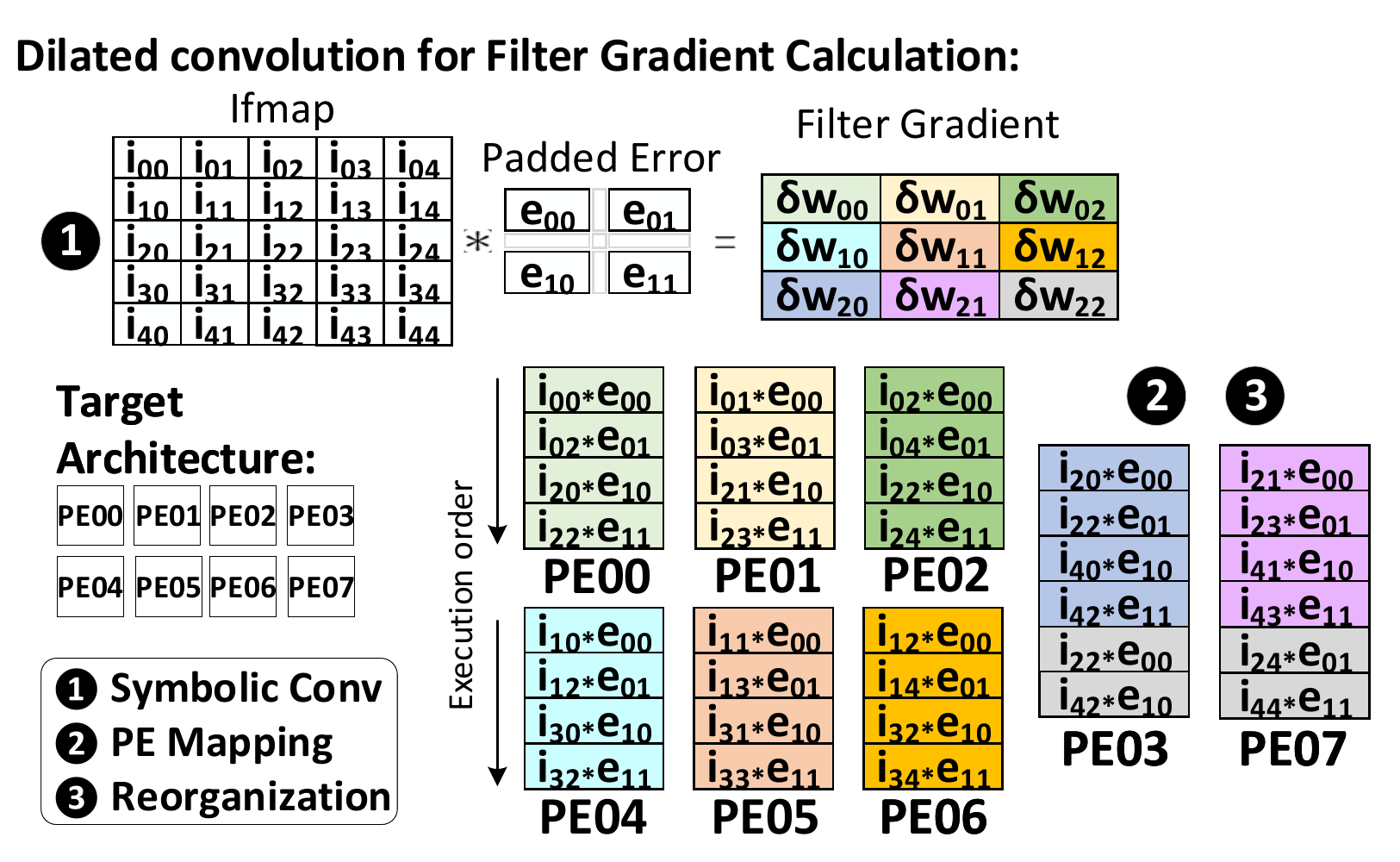}
    \caption{Dilated convolution using \mechanism{} to calculate the filter gradients in CNN training.}
    \label{fig:gradflow-fgrad}
\end{figure}

The \mechanism{} compiler also performs optimization techniques, called \emph{grouping} and \emph{expansion}, that allows to group high-dimension convolutions into a small PE array, or to expand small-dimension convolutions into a large PE array.

\vspace{1mm}
\subsubsection{Runtime.}
\vspace{1mm}
\label{sec:gradflow_igrad_runtime}
\label{sec:noc}
The \textbf{dataflow} in \mechanism{} leverages existing connections between vertically adjacent PEs and the on-chip multicast network present in spatial architectures. In this section, we describe data feeding and flow of the \textit{partial sums}, \textit{weights}, and \textit{error maps} through the PE array. Figure~\ref{fig:dataflow} summarizes the dataflow of the three data types through the PE array.

\emph{Partial sums} are accumulated locally and passed upward. Each filter-error product is added to a PE-local accumulation register. If a multiplication is the last one for a particular label (i.e., a color group) in one PE, the accumulated result is passed upward to the next PE in the same column.
In Figure~\ref{fig:gradflow-igrad}, the calculation of gradient element $\delta i_{22}$ needs three steps.  First, PE11 and PE01 compute $w_{00} \times e_{11}$ and $w_{20} \times e_{01}$, respectively. The results are stored in their internal accumulation registers.  Next, the PE's compute $w_{02} \times e_{10}$ and $w_{22} \times e_{00}$, adding the result to the accumulation register. Third, PE11 passes the value in its accumulation register to PE01, and PE01 adds the received value to its accumulation register. The result is $\delta i_{22}$, which is then stored into the off-chip memory.

\emph{Filter weights} are sequentially broadcast to all PEs and consumed every cycle. In Figure~\ref{fig:gradflow-igrad},
the first set of multiplications use $w_{00}$, which is used by all PEs ($w_{00} \times e_{00}$, $w_{00} \times e_{01}$, $w_{00} \times e_{10}$, $w_{00} \times e_{11}$). The next broadcast weight is $w_{10}$, and so on. %

\emph{Error matrix} elements are sequentially multicast to the PE array. Each PE maintains a list of multicast groups to which it is subscribed, and receives the error elements required. For example, in Figure~\ref{fig:gradflow-igrad}, PE00 receives the multicast groups $\{e_{00}, e_{01}\}$. Multicast groups are determined at compile time and loaded into each PE as part of an FSM.

\subsection{Dilated Convolution}
\label{sec:filter_gradients}

A dilated convolution is a direct convolution with a modified kernel (i.e., padded kernel) to match the desired output dimensions.
Without loss of generality, we use an example of dilated convolution that calculates the filter gradients in the CNN training algorithm. In this context, the input of the convolution is the ifmap from the forward pass, the filter corresponds to the padded error (the amount of padding depends on stride in the forward pass), and the output of the convolution are the calculated filter gradients.

\subsubsection{Compilation Time.}
\label{sec:gradflow_fgrad_compilation}
For a dilated convolution, \mechanism{}
performs computation scheduling and data mapping using a three-step process. For clarity, we walk through compilation using Figure~\ref{fig:gradflow-fgrad} which illustrates filter gradient calculation with a 5$\times$4 ifmap, 3$\times$3 filter, and stride 2 convolutional layer. 

\circled{1} \mechanism{} performs a symbolic convolution between the ifmap and the padded errors, determining the symbolic computations required to produce the filter gradients. During this step, the compiler forms groups that accumulate together to produce a single gradient element. In Figure~\ref{fig:gradflow-fgrad}, multiplications of the same color accumulate together.

\circled{2} \mechanism{} provisionally assigns the calculation of each filter gradient to one PE, eliminating inter-PE communications. Based on the necessity to parallelize channels in a filter, and to avoid potential slowdowns associated with large error maps, the compiler automatically reorganizes and re-distributes the compute schedule using assignment expansion, as explained in Section~\ref{sec:gradflow_fgrad_runtime}.

\circled{3} Finally, the compiler determines multicast groups for the ifmap for use during execution. In the next section, we describe the dataflow for the partial sums, error matrix, and ifmap.

\subsubsection{Runtime.}
\label{sec:gradflow_fgrad_runtime}
\mechanism{} uses a straightforward dataflow for calculating the filter gradients. Similar to the calculation of the input gradients, the calculation of the filter gradients adapts well to the underlying on-chip network in state-of-art spatial architectures. Figure~\ref{fig:dataflow} describes the three main dataflows.

\emph{Error matrix} elements are broadcast to each PE simultaneously. In Figure~\ref{fig:dataflow}, $e_{00}$ is used by all PEs in their first cycle, and is the first error to be broadcast.

The \emph{input matrix} is distributed to the PEs using a multicast pattern determined by the compiler. Figure~\ref{fig:dataflow} illustrates the input matrix multicast used for the convolution described in Figure~\ref{fig:gradflow-fgrad}. In Figure~\ref{fig:dataflow}, we see that each PE is part of at least four receive groups, corresponding to the number input matrix elements required in its computation schedule.

Finally, \emph{partial sums} are accumulated within the PE. Each PE is responsible for multiplying and accumulating all the data it receives, and for storing the resulting filter gradient into off-chip memory. With expansion, each PE follows a similar procedure: it multiplies and accumulates all the data mapped to it, and sends the final value to its vertical neighbor. The top PE, after performing all the local operations, accumulates any passed-in results, and writes the final gradient to memory.

\subsection{Memory Management}
\label{sec:memorymanagement}

We describe \mechanism{}'s data reuse using two concepts from ~\cite{Chen2017}. First, a \emph{PE set} is the subset of PEs used to run a 2D convolution. If the physical array is large enough, several PE sets can be mapped concurrently in the array. Second, a \emph{processing pass} is the contained, simultaneous execution of 2D convolutions in the PE array. In a single transpose convolution processing pass, each input element is read once from the global buffer, and the partial sums are stored back to the global buffer only once.

For a transpose or a dilated) convolution, \mechanism{} has three types of reuse: 1) it reuses the input values by storing them in the global buffer using them with different filters, 2) it reuses the filters by broadcasting and using them across multiple PEs, and 3) it accumulates the partial sums within the PE and across vertical PEs. The filters are streamed from DRAM directly to the PE registers, and the inputs and partial sums are stored in the global buffer for reuse between processing passes.

To map PE sets into a processing pass, \mechanism{} uses five parameters: $n$, $r$, $t$, $q$ and $p$. \mechanism{} fits $r \times t$ PE sets. Every $t$ PE sets share the same inputs with $t$ filters, and every $r$ PE sets that run on $r$ channels accumulate their partial sums within the PE array. Also, a processing pass can process $n$ inputs, $p$ filters and $q$ channels at the same time. These parameters depend on the size of the internal PE registers. \mechanism{} exhausts reuse opportunities of inputs and partial sum across different processing passes.

To optimize these parameters and allocate  global buffer space for inputs and partial sums, our compiler pass runs an optimization procedure that finds parameters that minimize energy consumption for a given hardware configuration.

\subsection{Hardware Architecture}
\label{sec:hw_architecture}

\mechanism{} targets spatial architectures similar to those described in Section~\ref{sec:background_spatial_architectures}.
We use Eyeriss~\cite{Chen2017} as the baseline architecture, and we incorporate changes to the on-chip network and PE array to support \mechanism{}.

\vspace{5pt}\noindent\textbf{On-Chip Network Requirements.} 
The baseline architecture uses four on-chip networks: 1) a filter broadcast network to send filter weights to the PE, 2) an ifmap multicast network to send a unique ifmap element to each PE (i.e., one multicast group per PE) 3) an ofmap network that delivers partial sums to the global buffer, and 4) a network of local unidirectional point-to-point links that transmit partial sums through PEs in a column.

\mechanism{} requires an expansion of the multicast network, so that each PE in the array can belong to several multicast groups. For example, in Figure~\ref{fig:gradflow-fgrad}, PE02 belongs to these four multicast groups: $\{i_{02}, i_{04}, i_{22}, i_{24}\}$. The multi-cast group $i_{02}$ is consists of PE00 and PE02, and likewise for other groups. To support this, we extend the original multicast network of Eyeriss~\cite{Chen2017}. To support an $R\times C$ array of PEs, Eyeriss has a vertical $Y$-bus consisting of $R$ horizontal $X$-buses. Each $X$-bus has a row ID, and each PE has a column ID. These IDs are reconfigurable, allowing different layers to map onto the same array.

We extend this network to have several row IDs per $X$-bus, and several column IDs per PE. For a $N \times N$ filter with stride $S$, the total number of row IDs that each $X$-bus needs to store is given by $\lceil \frac{N}{S} \rceil$. The number of bits needed by each row ID is $\lceil \left( \log_{2} 2N-S \right) \rceil$. $2N-S$ quantifies the total number of groups in a row. The equations to calculate the column ID requirements are exactly the same. We size the ID registers to support the largest layers in the CNN. For example, AlexNet requires five 5-bit row IDs per bus, while ResNet-50 requires four 4-bit row IDs per bus.

We estimate the area overhead of our NoC modifications by accounting for the additional logic gates and storage elements required to support the worst case CNN evaluated in this work. The extra IDs and comparison logic affect all the PE multicast controllers within the PE array. Our results show that the additional changes in the NoC introduce a 2.9\% area overhead in the PE array.

\mechanism{} also uses larger bandwidth to keep all PEs continuously utilized. Table~\ref{table:bandwidth} shows the maximum bus width required by \mechanism{} in the three networks to run at maximum throughput on all evaluated CNNs. First, \mechanism{} requires a 64+16 bits wide multicast global input network (GIN) for filters+ifmaps (forward pass), for errors+filters (input gradient calculation), and for ifmaps+errors (filter gradient calculation). Second, \mechanism{} requires a 64 bits wide global output network (GON) for ofmaps (forward pass), input gradients (input gradient calculation), and filter gradients (filter gradient calculation). Third, \mechanism{} requires a 64 bits wide local network (Local) for transmitting psums between vertical PEs.

\begin{table}[h]
    \centering
    \footnotesize
    \begin{tabular}{r  | c  c  c}
        \toprule
         & {\bf GIN} & {\bf GON} & {\bf Local} \\
        \midrule
        {\bf Eyeriss} & 64 + 16 bits & 64 bits & 64 bits \\
        {\bf \mechanism{}} & 80 + 32 bits & 64 bits  & 64 bits \\
        \bottomrule
    \end{tabular}
    \caption{Bus bit width of the multicast global input (GIN), global output (GON), and local (Local) networks.}
    \label{table:bandwidth}
\end{table}

We observe that \mechanism{} does not require additional bandwidth for GON and Local networks, and it requires 40\% more bandwidth for the GIN network.

\vspace{5pt}\noindent\textbf{PE Requirements.} Like a typical PE design, \mechanism{} needs an FSM to orchestrate loads and stores to registers, accumulations, stores to the global buffer, and communication with its neighboring PE. The compiler generates these FSMs. \mechanism{} accumulates in each PE a variable amount of partial sums before the PE sends the result to the above PE or to memory, and it needs to accumulate the corresponding partial sums together (e.g., the same colors needs to accumulated together in Figure~\ref{fig:gradflow-igrad}). This requires a slightly more complex FSM in the PE, compared to row-stationary dataflow. 

\vspace{5pt}\noindent\textbf{Memory Requirements.}
\mechanism{} does not require a different memory hierarchy than other spatial architecture accelerators. We use commodity DRAM chips and a highly banked global buffer. 

\section{SASiML: The Spatial Architecture Simulator}
\label{sec:sasim}

To evaluate \mechanism{}, we develop SASiML, a new cycle-accurate simulator that mimics the hardware of a spatial architecture. SASiML models all the components of the PEs, the network, and the memory hierarchy. Each component of SASiML can be fully microprogrammed, and all the latency and energy parameters are fully parametrizable. SASiML can estimate the latency and energy consumed by direct, transpose and dilated convolutions of a particular layer; many other metrics such as PE utilization and bandwidth can be measured. We also develop a new compiler that automatically generates the signals required by SASiML to execute a particular CNN layer. We \emph{open-source} both the simulator and the compiler to help enable the development of new dataflows and high-accuracy simulation environments for new spatial architectures and dataflows. This can be freely found at \url{https://github.com/CMU-SAFARI/sasiml}.

\subsection{The Simulator}

SASiML models the on-chip hardware of a spatial architecture and off-chip DRAM memory. SASiML contains architecture models for Eyeriss~\cite{Chen2017} and TPU~\cite{jouppi2017datacenter}. SASiML is extensible and fully programmable. The level of abstraction of SASiML is similar to RTL: we model a synchronous digital circuit in terms of the flow of digital signals (data) between hardware registers, and the logical operations performed on those signals. In addition to a timing simulator, SASiML is a functional simulator that propagates the input values through the PE array to get the output, which allows to validate that the implementation of the dataflow at microprogramming level is correctly implemented.

The simulator has three main components: (1) a PE array, each of which has a global buffer, local registers, pipelined multiply-and-accumulate unit, and input/output queues connected to neighbouring PEs (2) a network on chip that interconnects neighboring PEs and PEs to the global buffer, and (3) a highly banked global buffer (e.g., 27 banks in our evaluation in Section~\ref{sec:evaluation}). All components update their state at every clock cycle.

The basic organization of the simulator is simple: 1) the components connect together according to the specific design of the PEs and networks, 2) all components are controlled through input and output signals that are microprogrammed, and 3) all components update their state cycle by cycle. All components of SASiML are configurable, including memory sizes, network bandwidth, energy parameters. We support two variants of PEs, one tailored for convolutions (e.g., Eyeriss) and one for tailored for matrix multiplications (e.g., TPUs).

\subsection{The Compiler}

For simplifying the generation of the microprogramming control signals for SASiML, we implement a compiler. The inputs to the compiler are all the characteristics of the hardware and the CNN layers (e.g., feature map and filter dimensions). SASiML can perform inference and training with row-stationary, TPU, or \mechanism{} dataflows.

\subsection{Validation}

We validate SASiML by analyzing that the output values match the expected golden results, and that the timings and power consumption are similar to the results reported by a real chip Eyeriss accelerator~\cite{Chen2017}. We configure SASiML with the same row-stationary dataflow parameters and the same accelerator configuration as reported in~\cite{Chen2017}. Table~\ref{table:validation} shows the execution time, power, total size of global buffer accesses, and total size of all DRAM accesses for both Eyeriss and SASiML while running inference on AlexNet~\cite{krizhevsky2012}. We expect some variations because the Eyeriss paper~\cite{Chen2017} does not provide full detail about their exact procedure for measuring timing, and about their memory management mechanisms for convolutions with high filter/channel count that overflow the global buffer. We calculate the power based on the energy parameters for a 45nm technology node reported by Horowitz~\cite{horowitz2014}. There are two challenges for validating the power. First, the technology node of the Eyeriss chip is 65nm, not 45nm. We address this by scaling the energy consumption up by a factor of 1.4, based on estimations obtained from previous studies~\cite{Rodriguez2006}. Second, SASiML does not model the energy of many details that have a large influence in the energy consumption, such as the clock network, which consumes between ~33-45\% of the power~\cite{Chen2017}. We address this issue by using the Amdahl's law to estimate the total power consumption, so we are able to compare our results with the power consumed by the real chip~\cite{Chen2017}.

\begin{table}[h]
    \centering
    \begin{threeparttable}
    \footnotesize{}
    \setlength\tabcolsep{3.5pt} %
    \begin{tabular}{c|r|c|c|c|c|c|c}
        \toprule
        \multicolumn{1}{l}{} & \multicolumn{1}{l|}{} & {\bf CONV5} & {\bf CONV4} & {\bf CONV3} & {\bf CONV2} & {\bf CONV1} \\
        \midrule
        \multirow{4}{*}{\rotatebox[origin=c]{90}{{\bf SASiML}}} &{\bf Exec. Time} & 12.5ms & 18.8ms & 25ms & 39.5ms & 15.2ms \\
         \cline{2-7}
         &{\bf Power} & 207mW & * & * & * & 273mW \\
        \cline{2-7}
         &{\bf GB acc.} & 23.8MB & 35.6MB & 66MB & 74MB & 16.8MB \\
         \cline{2-7}
        
         &{\bf DRAM acc.} & 1.5MB & 2.1MB & 2.6MB & 4.11MB & 3.6MB \\
\hline
        \multirow{4}{*}{\rotatebox[origin=c]{90}{{\bf Eyeriss}}}&{\bf Exec. Time} & 11ms & 16ms & 21.8ms & 39.2ms & 16.5ms \\
        \cline{2-7}
         &{\bf Power} & 236mW & 235mW & 266mW & 288mW & 332mW \\
        \cline{2-7}
        &{\bf GB acc.} & 24.9MB & 37.4MB & 50.2MB & 77.6MB & 18.5MB \\

        \cline{2-7}
        &{\bf DRAM acc.} & 1.3MB & 2.1MB & 3.0MB & 4.0MB & 5.0MB \\
        \bottomrule
    \end{tabular}
    \begin{tablenotes}
     \item[*] Eyeriss~\cite{eyerissv2} does not report the detailed power breakdown of these layers, so it is not possible to cross-verify these particular results.
    \end{tablenotes}
    \end{threeparttable}
    \caption{Comparison of execution time, global buffer (GB) accesses and DRAM accesses of SASiML and Eyeriss~\cite{Chen2017}.}
    \label{table:validation}

\end{table}

We observe that the results of SASiML are similar to the real chip Eyeriss measurements, and follow the same trends across layers. We make three key observations. First, the reported SASiML execution time is within 0.07\% to 10\% of the real Eyeriss accelerator. Second, the amount of data accessed in memory (GB and DRAM) by SASiML has a deviation of 0\% to 24\% from real measurements. Third, the power consumption reported by SASiML shows a good approximation, and the results are relatively accurate, despite the fact we could not model many details that are missing in the real Eyeriss chip paper.

We conclude that SASiML is an cycle accurate simulator that allows to model different spatial arrays with different NoCs and PE configurations at a microprogramming level of detail, which enables to functionally verify the correctness of dataflows and its implementation.

\section{Evaluation}
\label{sec:evaluation}

We evaluate transpose and dilated convolutions using workloads that contain both types of convolutions: CNN training (Section~\ref{sec:evaluation_CNN}) and GAN training (Section~\ref{sec:evaluation_GANs}).

\subsection{Experimental Setup}
\label{sec:methodology}

We use the SASiML simulator and the SASiML compiler (Section~\ref{sec:sasim}) to evaluate \mechanism{}. We model the energy of the accelerator with values obtained from a 45nm process~\cite{horowitz2014}. We model  DRAM energy using  DRAMPower~\cite{chandrasekar2012drampower}. We compare \mechanism{} to the row-stationary (RS) dataflow~\cite{Chen2017} used in Eyeriss and to a lowering-based convolutional dataflow used in TPUs~\cite{jouppi2017datacenter,jouppi2021ten}. Table~\ref{table:eyeriss_conf} shows the configuration of the target architecture used in evaluation. We chose an array of 13$\times$15 PE elements, matching prior work and tuned with RS and TPU dataflows to fit the dimensions of the evaluated layers.

\begin{table}[h]
    \centering
    \footnotesize
    \begin{tabular}{ r | l }
        \toprule
        {\bf PE Array}  & 13 x 15 PEs \\
        {\bf PE Array Clock}  & 200 MHz \\
        {\bf PE Register File (ifmap, filter, psum)}  & 75, 224, 24 \\
        {\bf PE Register Latency}  & 1 cycle \\
        {\bf Global Buffer}  & 108KB / 27 banks \\
        {\bf DRAM}  & 4GB DDR4 1866MHz \\
        {\bf Clock Gating}  &  Zero Operations\\
        {\bf Multiplier/Accumulator}  & 2-stage/1-stage \\
        {\bf I/O Queues}  & 8 entries \\
        {\bf On-chip Network Latency}  & 1 cycle \\
        \bottomrule
    \end{tabular}
    \caption{Configuration of the base CNN accelerator.}
    \label{table:eyeriss_conf}
\end{table}

We implement a clock-gating mechanism that activates when the PE receives a zero value \cite{Chen2017}. This is included in all our baselines. The NoC of Eyeriss and \mechanism{} are similar, implementing dedicated networks for each data type. We use the on-chip networks described in Table~\ref{table:bandwidth}. The TPU uses a much simpler NoC with only two uni-directional connections between neighbour PEs (for propagating input and filter values), while the partial sums are accumulated locally. We evaluate CNN training in Section~\ref{sec:evaluation_CNN} and GANs in Section~\ref{sec:evaluation_GANs}.

To estimate the execution time of the end-to-end CNN training algorithm (i.e., execution time of all layers), we first profile the evaluated models in GPU and CPU to get the average breakdown of the execution time per layer, and we apply the Amdahl's law to calculate the expected total performance gains.

\subsubsection{Optimizing CNN Training for \mechanism{}}
\label{sec:optimization}

To get the maximum benefit from \mechanism{} on CNN training, we need to replace pooling layers with larger strides when possible. %
Prior work demonstrates that pooling can be replaced by a convolutional layer with increased stride \emph{without loss in accuracy} \cite{springenberg2014striving}. The authors show that for the tested CNNs, when they replace pooling with a convolutional layer with 2-stride, there is no accuracy loss. We corroborate and extend these results with experiments of our own on six larger, more recent CNNs. We train two variants of each CNN topology: one with pooling layers and one with pooling layers replaced with larger stride. We use the CIFAR-10~\cite{cifar10} and ImageNet~\cite{russakovsky2015imagenet} training and test datasets, and retain the default learning hyper-parameters given in \cite{cifar10models, pytorch}.

Table~\ref{table:accuracy_network} summarizes our results. We observe that using a larger stride (Stride) instead of pooling layers marginally reduces accuracy ($<$2\%), and in some cases, improves accuracy. This can be an acceptable trade-off in some applications, given the performance advantages.

\begin{table}[h]
    \centering
    \footnotesize
    \setlength\tabcolsep{2pt} %
    \begin{tabular}{ r | c | c | c | c | c | c }
                \toprule
         \multicolumn{1}{c|}{}&\multicolumn{3}{c|}{{\bf CIFAR-10}}& \multicolumn{3}{c}{{\bf ImageNet}} \\

        {\bf CNN} & {\bf Original} & {\bf Stride} & {\bf Diff.} & {\bf Original} & {\bf Stride} & {\bf Diff.} \\
        \midrule
        ResNet-18 \cite{resnet} & 94.6\% & 94.2\% & -0.4\% & 69.6\% & 69.5\% & -0.1\% \\
        ResNet-101 \cite{resnet} & 94.6\% & 93.7\% & -0.9\% & 77.6\% & 76.9\% & -0.7\%\\
        DenseNet-201 \cite{densenet} & 94.0\% & 93.7\% & -0.3\% & 78.6\% & 76.8\% & -1.8\%\\
        VGG-19 \cite{vgg} & 92.5\% & 92.1\% & -0.4 & 74.5\% & 74.6\% & +0.1\%\\
        MobileNet-v2 \cite{mobilenet} & 90.7\% & 90.7\% & +0.0\% & 74.7\% & 73.14\% & -1.56\%\\
        \bottomrule
  
    \end{tabular}
    \caption{Accuracy comparison of CNNs that downsample using pooling layers (original) versus a larger stride (Stride).} 
    \label{table:accuracy_network}
\end{table}

\subsection{CNN Training Evaluation}
\label{sec:evaluation_CNN}

Table~\ref{table:evaluated_layers} details characteristics of 8 sample layers that we evaluate from six representative and widely-used CNNs, namely AlexNet~\cite{krizhevsky2012}, ResNet-50~\cite{resnet}, Shufflenet~\cite{zhang2018shufflenet}, Inception~\cite{szegedy2015going}, Xception~\cite{chollet2017xception}, and MobileNet~\cite{mobilenet}.

Our complete evaluation tested 72 layers in total. These layer topologies and networks encompass a most of the layers used in popular networks, and include recent winning topologies of the ILSVRC competitions \cite{russakovsky2015imagenet}. We use a batch size of four in our evaluations. We also evaluate the variant of each layer that includes the larger stride optimization described in Section~\ref{sec:optimization}. We denote these layers with a suffix of \texttt{opt}.

\begin{table}[h]
    \centering
    \footnotesize
    \setlength\tabcolsep{3pt} %
    \begin{tabular}{ r | c | c | c | c | c | c | c }
        \toprule
        {\bf CNN} & {\bf Layer\#} & {\bf IFM} & {\bf OFM} & {\bf Filter} & {\bf \# Filts} & {\bf Str.} & {\bf Opt.}\\
        \midrule
        AlexNet & CONV1 & 3x224x224 & 55x55 & 11x11 & 64 & 4 & Yes\\
        AlexNet & CONV2 & 64x31x31 & 27x27 & 5x5 & 192 & 1 & Yes\\
        ResNet-50 & CONV3 & 128x57x57 & 28x28 & 3x3 & 128 & 2 & No\\
        ShuffleNet & CONV2 & 58x57x57 & 28x28 & 3x3 & 58 & 2 & No\\
        ShuffleNet & CONV5 & 232x7x7 & 7x7 & 1x1 & 232 & 1 & No\\
        Inception & CONV3 & 192x17x17 & 8x8 & 3x3 & 320 & 2 & No\\
        Xception & CONV3 & 728x29x29 & 14x14 & 3x3 & 1 & 2 & No\\
        MobileNet & CONV5 & 512x15x15 & 7x7 & 3x3 & 1 & 2 & No\\
        \bottomrule
    \end{tabular}
    \caption{Eight of the 72 evaluated layers from three CNNs.} 
    \label{table:evaluated_layers}
\end{table}

We train using 16 bits instead of the 32 bits used in typical training algorithms. A previous work~\cite{kalamkar2019study}  demonstrates, training with BFLOAT16 can achieve the same accuracy as training with FP32.

\subsubsection{Performance results.}
\label{sec:performance}

\begin{figure*}[h] 
\centering
    \includegraphics[width=0.75\linewidth]{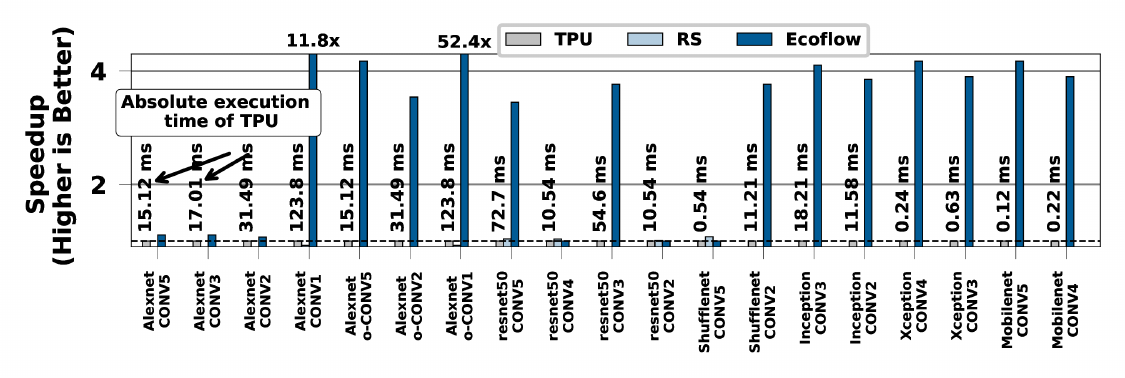}
    \caption{Speedup of input gradient calculation, normalized to the TPU dataflow, and absolute TPU execution time.}
    \label{fig:performance_input}
\end{figure*}

\begin{figure*}[h] 
\centering
    \includegraphics[width=0.75\linewidth]{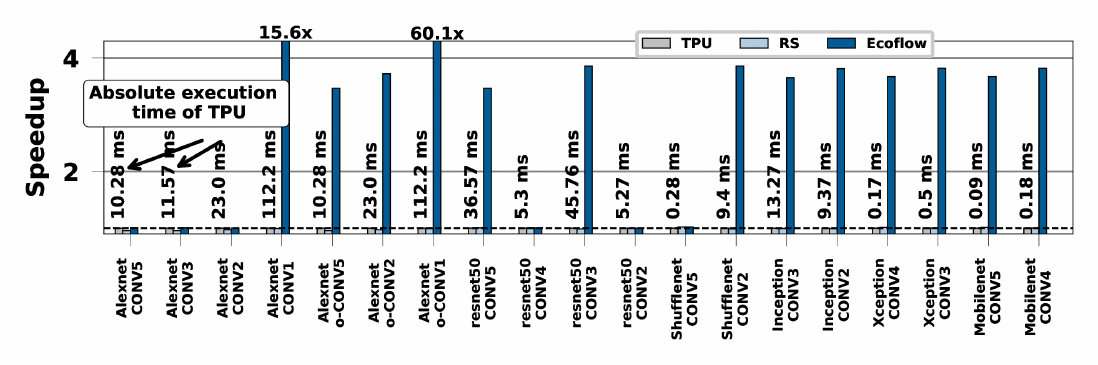}
    \caption{Speedup of the filter gradient calculation, normalized to the TPU dataflow.}
    \label{fig:performance_filter}
\end{figure*}

Figure~\ref{fig:performance_input} shows the speedup of input gradient calculation through each layer in TPU, RS and \mechanism{} dataflows, normalized to TPU. Similarly,  Figure~\ref{fig:performance_filter} shows the speedup of the filter gradient calculation for the three dataflows. The numbers on top of the TPU bars indicate the absolute execution time of TPU in \emph{milliseconds}. The layers starting with the letter "o" (e.g., Alexnet o-CONV1) are the optimized versions of the layers (Section~\ref{sec:optimization}).

We make two main observations. First, the speedup of \mechanism{} for calculating the input gradients compared to TPU and RS is very high for strides larger than 1. As shown in the figure, the speedup is close to 4x for stride 2 (e.g, resnet50 CONV3), 11x for stride 4 (Alexnet CONV1), and 52x for stride 8 (Alexnet opt CONV1). For stride 1, the speedup is from 0\% (e.g., resnet50 CONV2) to 10\% (Alexnet CONV3). Second, the speedup of \mechanism{} for calculating the filter gradients compared to TPU and RS is also very large for stride larger than 1. The speedup is more than 3x for stride 2 (e.g., resnet50 CONV3), 15.6x for stride 4 (Alexnet CONV1), and 60.1x for stride 8 (Alexnet o-CONV1). We conclude that \mechanism{} performs the backward pass much more efficiently than RS and TPUs, especially for strides larger than 1.

\begin{figure*}[h] 
\centering
    \includegraphics[width=0.9\linewidth,trim=4 4 4 4, clip]{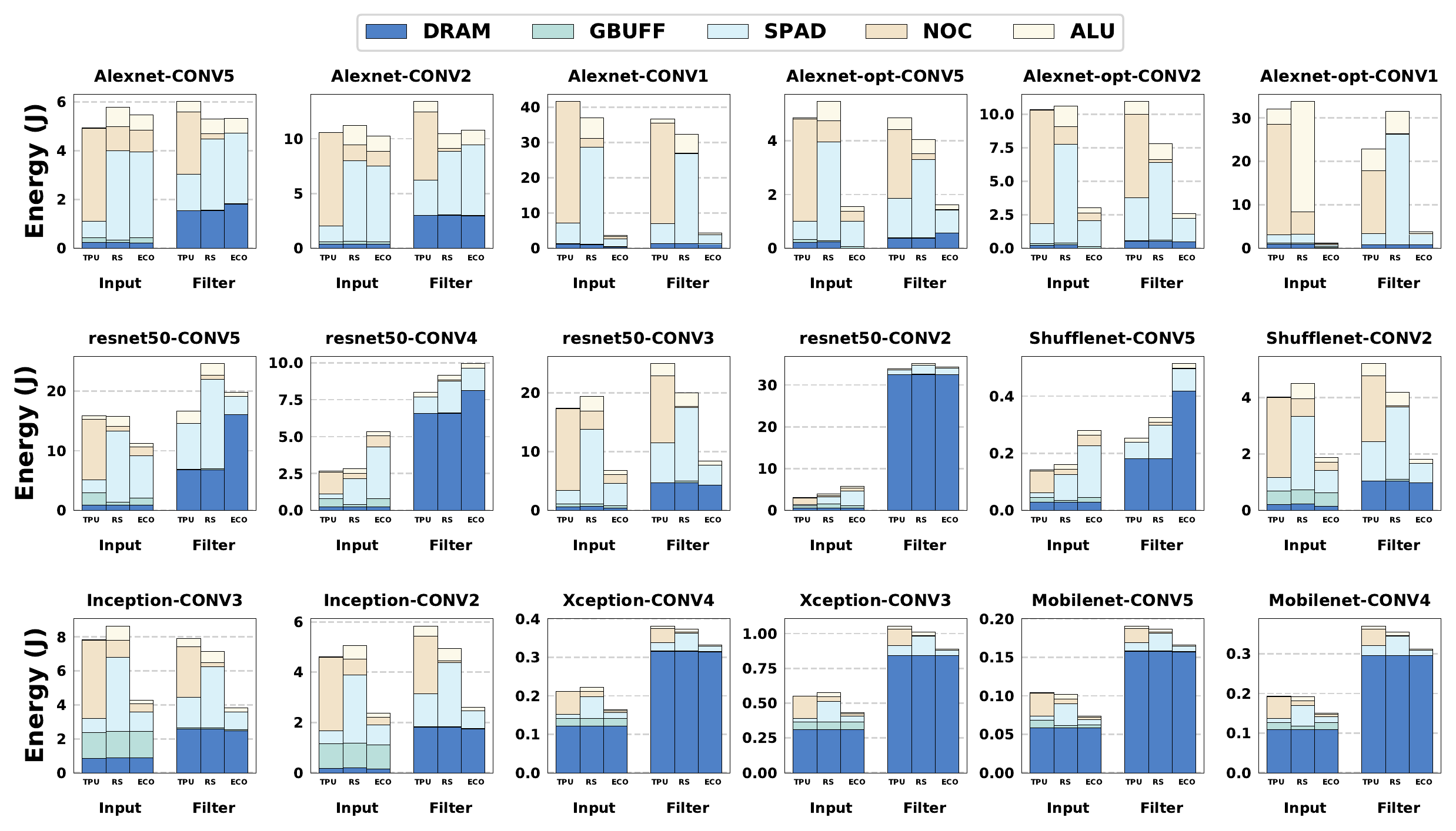}
    \caption{Energy consumption of the evaluated layers.}
    \label{fig:energy_results}
\end{figure*}

Table~\ref{table:entireNetwork} shows the speedup of the evaluated end-to-end CNN networks.

\begin{table}[h]
    \centering
    \footnotesize
    \setlength\tabcolsep{3pt}
    \begin{tabular}{r|c|c|c|c|c|c}
         \toprule
        \multicolumn{1}{c}{ } & \multicolumn{3}{|c|}{ {{\bf Speedup}}} & \multicolumn{3}{c}{{{\bf Energy savings}}} \\

        \cmidrule{2-7} %
        & {\bf TPU} & {\bf Eyeriss} & {\bf \mechanism{}} & {{\bf TPU}} & {{\bf Eyeriss}} & {{\bf \mechanism{}}} \\  %
    \midrule
    Alexnet & 1 & 0.94 &  1.83 & {1} & {0.97} & {1.38} \\  %
    ResNet-50 & 1 & 0.99 & 1.07 & {1} & {1.02} & {1.06}\\  %
    ShuffleNet & 1 & 0.98 & 1.08 & {1} & {1} & {1.07}\\  %
    Inception & {1} & {1.01} & {1.08} & {1} & {0.99} & {1.08}\\  %
    Xception & {1} & {1.01} & {1.11} & {1} & {1.00} & {1.10}\\  %
    Mobilenet & {1} & {1.01} & {1.09} & {1} & {1.00} & {1.08}\\  %
    \bottomrule
    \end{tabular}
    \caption{Speedup {and energy savings} of end-to-end CNN training of convolutional layers in different architectures, normalized to TPU {(larger is better)}.}
    \label{table:entireNetwork}
\end{table}

We make two observations. First, Alexnet greatly benefits from \mechanism{}, because more than 80\% of the execution time is dedicated to execute convolution layers following by pooling layers, or convolutional layer with stride larger than one. Second, ResNet-50, ShuffleNet, {Inception, Xception and Mobilenet} have smaller benefits because many of their convolutional layers have stride 1. We conclude that \mechanism{} has very significant end-to-end benefits in networks that use strides in convolutional or pooling layers. Notice that other modern networks with larger strides, like EfficientNet~\cite{tan2019efficientnet}, would also greatly benefit from \mechanism{}.

\subsubsection{Energy Results}

In this section, we evaluate the energy consumption of \mechanism{}. PEs are clock gated when idle, and all other parameters are defined in Table~\ref{table:eyeriss_conf}.

Figure~\ref{fig:energy_results} shows the energy comparison of TPU, RS and \mechanism{} for the input gradient and filter gradient calculation. The breakdown of the energy includes DRAM (DRAM), global buffer (GBUFF), internal scratchpad memories (SPAD), the multipliers and the adders (ALU), and all on-chip networks (NoC). We make {four} main observations. First, the energy consumption of \mechanism{} is much lower than TPU and RS for strides larger than 1. For example, the maximum energy savings of \mechanism{} is 26x for Alexnet-opt-CONV1 compared to TPU. For the filter gradients, \mechanism{} saves up to 8.3x energy. Second, the energy savings of \mechanism{} are coming mainly from SPAD and NoC, whereas the energy consumed by DRAM is maintained. Third, for some layers with stride 1, \mechanism{} consumes more energy than TPU and RS, caused by an increased DRAM energy consumption. {Four, the energy of the filter gradient calculation is dominated by DRAM in some layers, e.g., resnet50-CONV4, resnet50-CONV2, since the errors  in these layers have little reuse and are memory bound. This happens when the kernel size is small}. \mechanism{} is most energy efficient for layers that have stride and kernel larger than one.

\subsection{GAN Evaluation}
\label{sec:evaluation_GANs}

In this section, we evaluate GAN convolutional layers executed in the spatial architecture described in Table~\ref{table:eyeriss_conf}. We compare \mechanism{} to GANAX~\cite{ganax}, a hardware GAN accelerator that optimizes the execution of GANs by avoiding unnecessary zero computations. The key idea introduced by GANAX is to identify repeated patterns in the GAN computation and create different microprograms to execute each of this patterns. GANAX requires significant changes over an Eyeriss architecture, a new SIMD-MIMD execution model, a new ISA, a new global buffer to store instructions, and decoupling of the PEs into execution units and access units.

Table~\ref{table:evaluated_GAN_layers} shows the properties of the evaluated GAN layers. The layers are used by two representative GANs, namely CycleGAN~\cite{cyclegan}, and pix2pix~\cite{pix2pix}. The layers of the discriminator (Disc) are regular convolutional layers, and the layers of the generator (Gen) are transposed convolutions. \mechanism{} accelerates the backward pass of the discriminator and the forward pass of the generator.

\begin{table}[h]
    \centering
    \footnotesize
    \setlength\tabcolsep{1pt} %
    \begin{tabular}{ r | c | c | c | c | c | c }
        \toprule
        {\bf CNN} & {\bf Layer\#} & {\bf IFM} & {\bf OFM} & {\bf Filter} & {\bf \# Filts} & {\bf Str.} \\
        \midrule
        CycleGAN & Disc-CONV3 & 64x114x114 & 56x56 & 4x4 & 128 & 2 \\
        CycleGAN & Gen-TCONV1 & 256x56x56 & 113x113 & 3x3 & 128 & 2 \\
        pix2pix  & Disc-CONV6 & 128x130x130 & 64x64 & 4x4 & 256 & 2 \\
        pix2pix  & Gen-TCONV41 & 512x64x64 & 130x130 & 4x4 & 128 & 2 \\
        \bottomrule
    \end{tabular}
    \caption{Evaluated layers from two widely-used GANs.} 
        \label{table:evaluated_GAN_layers}
\end{table}

\subsubsection{Performance Results}

\begin{figure}[h] 
\centering
    \includegraphics[width=0.99\linewidth]{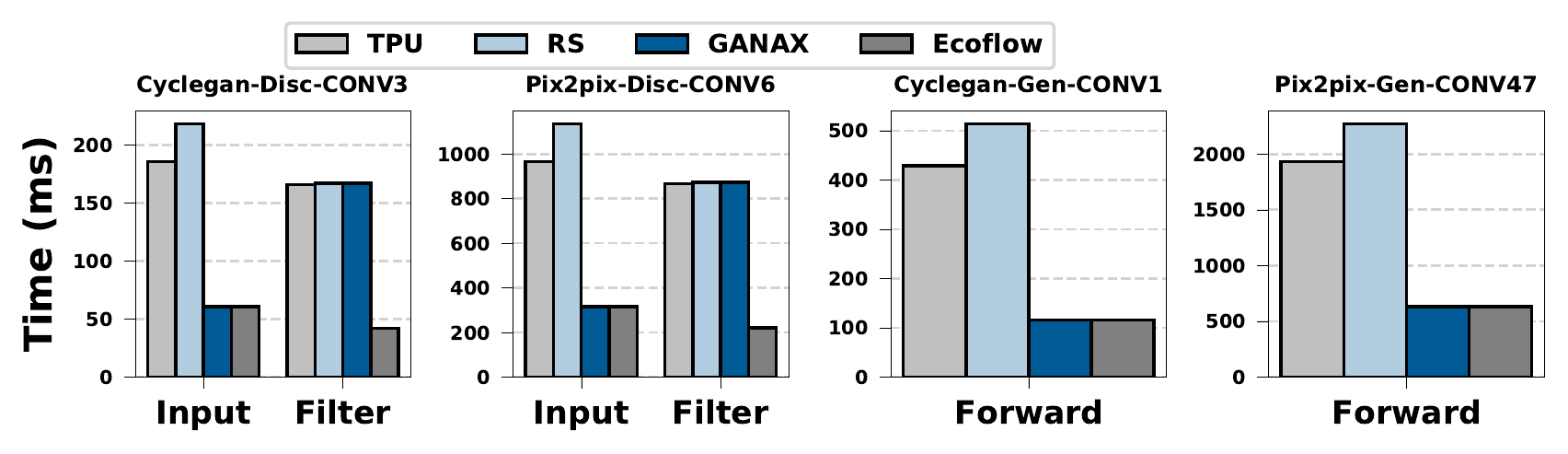}
    \caption{Execution time of the evaluated GAN layers.}
    \label{fig:performance_gans}
\end{figure}

{Figure~\ref{fig:performance_gans} shows the speedup of the backward (Input, Filter) and the forward passes of selected GAN layers, for RS, TPU, GANAX, and \mechanism{} dataflows, normalized to RS. We make two observations. First, \mechanism{} performs on the order of 4x better than RS and TPU. Because GANs use strides larger than 1 instead of pooling layers, \mechanism{} accelerates most convolutional layers. Second, \mechanism{} performs 3-4x times better than GANAX in the filter gradient calculations, because GANAX does not provide a dataflow to accelerate gradient calculation. However, GANAX performs very similar to \mechanism{} in the forward pass of the generative layers, and in the calculation of the input gradients.}

Table~\ref{table:entireGANNetwork} shows the speedup of the evaluated end-to-end GAN networks.

\begin{table}[h!]
    \centering
    \footnotesize
    \setlength\tabcolsep{1pt}
    \begin{tabular}{r|c|c|c|c|c|c|c|c}
    \toprule
    \multicolumn{1}{c|}{ } & \multicolumn{4}{c|}{ {{\bf Speedup}}} & \multicolumn{4}{c}{{{\bf Energy savings}}} \\
        \cmidrule{2-9} %
        & {\bf TPU} & {\bf Eye.}& {\bf GANAX} & {{\bf \mechanism{}}}& {{\bf TPU}} & {{\bf Eye.}} & {\bf GANAX}& {{\bf \mechanism{}}}  \\  %
    \midrule
    pix2pix & 1 & 0.95 & 1.34 & 1.39 & {1} & {0.93} & 1.11 & {1.29} \\  %
    CGAN & 1 & 0.94 & 1.37 & 1.42 & {1} & {1.04} & 1.32 &{1.37}\\  %
    \bottomrule
    \end{tabular}
    \caption{Speedup {and energy savings (higher is better)} of end-to-end training of two GANs, normalized to TPU.}

    \label{table:entireGANNetwork}
\end{table}

We make the key observation that \mechanism{} has large benefits in end-to-end training of GAN networks. The training performance of \mechanism{} outperforms even specialized GAN architectures like GANAX, because \mechanism{} can accelerate filter gradient calculations.

\subsubsection{Energy Results}
\begin{figure}[h] 
\centering
    \includegraphics[width=0.9\linewidth]{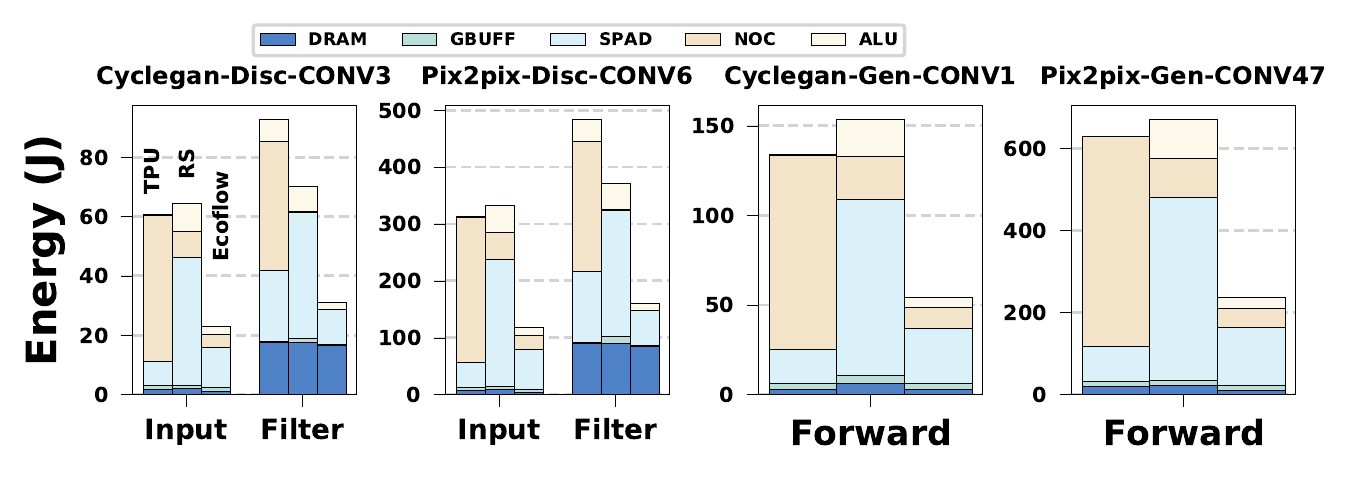}
    \caption{Energy breakdown of the evaluated GAN layers. }
    \label{fig:energy_gans}
\end{figure}

{Figure~\ref{fig:energy_gans}} shows the energy breakdown of the backward (Input,Filter) and the forward passes of selected GAN layers, for TPU, RS and \mechanism{} dataflows, in absolute values. {We could not compare to GANAX because some implementation details are missing in the paper (e.g., data reuse in each memory).}%

We make two main observations. First, the energy consumption of \mechanism{} is much lower than the energy consumption of TPU and RS. For example, for the cyclegan-disc-CONV3 layer the energy savings of \mechanism{} are in the order of 4x compared to TPU and RS. Second, similar to the results in CNN training (Section~\ref{sec:evaluation_CNN}), the energy savings of \mechanism{} are coming from reducing the energy in the SPADs, NOC and ALUs, whereas the DRAM energy consumption is very similar in all dataflows. 

A key property of GANs is that they use larger strides instead of pooling layers, so most of the layers of state-of-the-art GANs benefit from \mechanism{}.

\section{Related Work}
\label{sec:related_work}

To our knowledge, this is the first work to design efficient dataflows to perform transpose and dilated convolutions on low-power CNN inference accelerators. We have already extensively compared \mechanism{} to the Google  TPU~\cite{jouppi2017datacenter,jouppi2021ten}, Eyeriss~\cite{Chen2017} and GANAX~\cite{ganax}. In this section, we describe other related works.

\vspace{5pt}\noindent\textbf{Specialized Inference Accelerators}. Most existing specialized CNN accelerators are optimized for direct convolutions commonly used on CNN inference (e.g. Eyeriss~\cite{Chen2017}, DaDiannao~\cite{Chen2014}, Tetris \cite{tetris}, and Minerva \cite{minerva}). WaveCore~\cite{wavecore} and Google's TPUv2~\cite{jouppi2017datacenter} support CNN training, but suffer from challenges highlighted in Section~\ref{sec:challenges}. \mechanism{} solves these issues, while introducing minimal changes to the CNN inference accelerator architecture.

\vspace{5pt}\noindent\textbf{Specialized Training Accelerators}. Cambricon-Q~\cite{zhao2021cambricon} proposes a hybrid architecture consisting of an ASIC acceleration core and a near-data-processing (NDP) engine with the goal of improving the efficiency of statistic-based quantization. Equinox~\cite{drumond2021equinox} proposes a custom inference accelerator that has the main goal of interleaving training during idle inference cycles. FPRaker~\cite{awad2021fpraker} proposes a processing element that can perform MAC operations concurrently to accelerate DNN training. Unlike these works, \mechanism{} targets a different problem, which is the inefficiency of convolutional dataflows used in DNN training and other DNN workloads. 

\vspace{5pt}\noindent\textbf{Sparse Accelerators}. Sparse accelerators~\cite{han2015deep, eie, cambriconx, scnn, aimar2018nullhop, mao2017exploring, page2017sparcnet,nakahara2019fpga,jang2021sparsity,lu2021distilling,gondimalla2019sparten} address the inefficiencies caused by zeros contained in sparse matrices, which is a fundamentally different problem than padding introduced by transpose and dilated convolutions. \mechanism{} can be incorporated to these accelerators to obtain aggregated benefits.

\vspace{5pt}\noindent\textbf{GAN Accelerators.} {Prior works focus on accelerating GANs by performing transposed convolutions on new memory technologies \cite{zara,mao2018lergan}, FPGAs~\cite{yazdanbakhsh2018flexigan}, and significantly modified spatial architectures \cite{ganax, im2019dt,im2020dt}.}
Our work is unique in that 1) it focuses on both transposed and dilated convolutions, 2) it requires fewer hardware changes, 3) it proposes a multicast network that is able to effectively distribute the input data into the corresponding PEs, and 4) it evaluates GAN training and CNN training.

\vspace{5pt}\noindent\textbf{{Winograd and Frequency-Domain Algorithms}}. Winograd is an alternative algorithm to perform matrix multiplications~\cite{kim2019efficient,shi2018sparse,liu2018efficient} that reduces the number of computations in CNNs via a series of data transformations. GradFlow, however,  targets the orthogonal problem of the zero padding introduced by the training algorithm to upscale and back-propagate the errors through the network.
Frequency domain backpropagation~\cite{ko2017design} replaces convolutions with simple point-wise multiplications, which avoids the inefficiencies of transposed and dilated convolutions. However, this approach requires computationally-intensive Fast Fourier Transforms (FFTs) and Inverse FFTs (IFFTs) at the boundary of every layer, and it requires a larger memory footprint.

\vspace{5pt}\noindent\textbf{Other Algorithms.} Direct convolutions~\cite{zhang2018high,georganas2018anatomy} can avoid zero padding in the backward pass of some layers that meet some specific and restricted parameters. In contrast, \mechanism{} is a general dataflow that can apply to the backward pass of any convolutional layer.

\vspace{5pt}\noindent\textbf{Other Techniques to Improve the Efficiency of DNN Workloads.} There are other techniques to improve performance and reduce energy consumption in DNN workloads~\cite{nguyen2019st,nguyen2018approximate,tu2018rana,koppula2019eden,chandramoorthy2019resilient,boroumand2021google,boroumand2021mitigating,kwon2021heterogeneous, jeong2021rasa,munoz2021novel,rashidi2021themis,jeong2021union,choi2020prema,chatarasi2021marvel,alwani2016fused,gao2019tangram,shen2017maximizing,jafri2020refresh,salami2020experimental,imani2019floatpim,jiang2020cimat,zhang2020robust,cho2021accelerating,long2020q,shafiee2016isaac,chi2016prime,song2017pipelayer,tetris,fowers2018,sharma2018bit,ankit2019puma,gokmen2017training,cheng2017time,azarkhish2017neurostream,kwon2019tensordimm,pimtraining,yang2019sparse,angizi2018cmp,wang2018snrram,chen2018regan,fan2017energy,ji2018recom,deng2019tie,chen2018design}. For example, EDEN~\cite{koppula2019eden} reduces energy consumption by reducing the timing parameters and the voltage of DRAM, while \cite{chandramoorthy2019resilient} improves energy efficiency by reducing the voltage of SRAM in a DNN accelerator.
Mensa~\cite{boroumand2021google,boroumand2021mitigating} tackle the problem of heterogeneity in ML workloads by considering several aspects (e.g., off-chip memory, on-chip buffers, compute-centric vs. data-centric acceleration, dataflow, etc.) to propose a family of accelerators where each accelerator tackles a different ML workload or layer. 
FloatPIM~\cite{imani2019floatpim} is a Processing-in-Memory (PIM)~\cite{mutlu2020modern,ghose2019processing} approach that  natively supports floating-point representation in resistive memories for CNN training workloads.
Unlike \mechanism{}, these approaches do not fundamentally re-design the dataflow of dilated and transposed convolutions to avoid inefficiencies in low-cost accelerators with limited hardware resources.

\section{Conclusion}

In this work, we aim to accelerate transpose and dilated convolutions in energy-efficient spatial architectures designed for CNN inference. We observe that a main source of inefficiencies of state-of-the-art  CNN inference accelerators when executing transpose and dilated convolutions is the large amount of required zero padding, which diminishes the overall energy efficiency and performance. 

To address this issue, we propose \mechanism{}, a new set of mapping and dataflows for transpose and dilated convolutions. \mechanism{} eliminates zero-padding by meticulously orchestrating the scheduling, dataflow, and data mapping to fit the characteristics of the target CNN inference accelerator. We show that, by introducing minimal changes to the CNN inference hardware, \mechanism{} can significantly improve the energy efficiency and performance of common CNN training workloads. We conclude that \mechanism{} enables commonly-used low-power CNN inference accelerators to efficiently perform CNN training, GAN training and other workloads that use transpose and dilated convolutions, with minimal hardware changes. We hope \mechanism{} inspires future works on ML acceleration that take into account such important training workloads.

\section*{Acknowledgments} 
We thank the anonymous reviewers of ISCA\textquotesingle21/20/19, ASPLOS\textquotesingle20, and MICRO\textquotesingle20/19  for feedback, and the SAFARI Research Group members for valuable feedback and the stimulating intellectual environment they provide. We thank Taha Shahroodi for his feedback on earlier versions of this paper.
We acknowledge the generous gifts provided by our industrial partners: Google, Huawei, Intel, Microsoft, VMware, and the Semiconductor Research Corporation.

\bibliographystyle{IEEEtran}
\bibliography{refs}

\newpage
\begin{IEEEbiography}[{\includegraphics[width=1in,height=1.25in,clip,keepaspectratio]{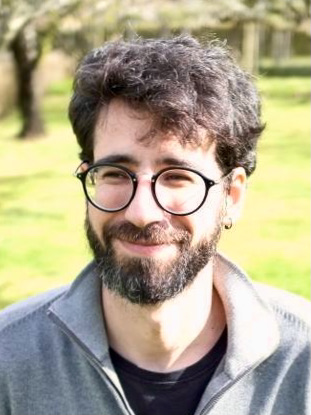}}]{Lois Orosa}
is a senior researcher at SAFARI Research group @ ETH Zürich, Switzerland. He received his BS and MS degrees in Telecommunication Engineering from the University of Vigo, Spain, his PhD degree from the University of Santiago de Compostela, Spain, and he held a postDoc position in the University of Campinas, Brazil. He was a visiting researcher at multiple companies (IBM, Recore Systems, Xilinx and Huawei) and universities (UIUC and Universidade Nova de Lisboa). His current research interests are in computer architecture, hardware security, reliability, memory systems, and machine learning (ML) accelerators. For more information,
please see his webpage at \url{https://loisorosa.github.io/}.
\end{IEEEbiography}

\begin{IEEEbiography}[{\includegraphics[width=1in,height=1.25in,clip,keepaspectratio]{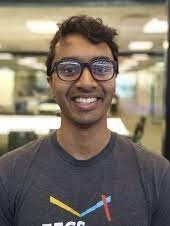}}]{Skanda Koppula}
is currently a research engineer at DeepMind. Previous to this, he worked at ETH Zürich in the SAFARI research group on memory systems, machine learning acceleration, and computer architecture. He completed his MEng and BSc from MIT in 2018.
\end{IEEEbiography}

\begin{IEEEbiography}[{\includegraphics[width=1in,height=1.25in,clip,keepaspectratio]{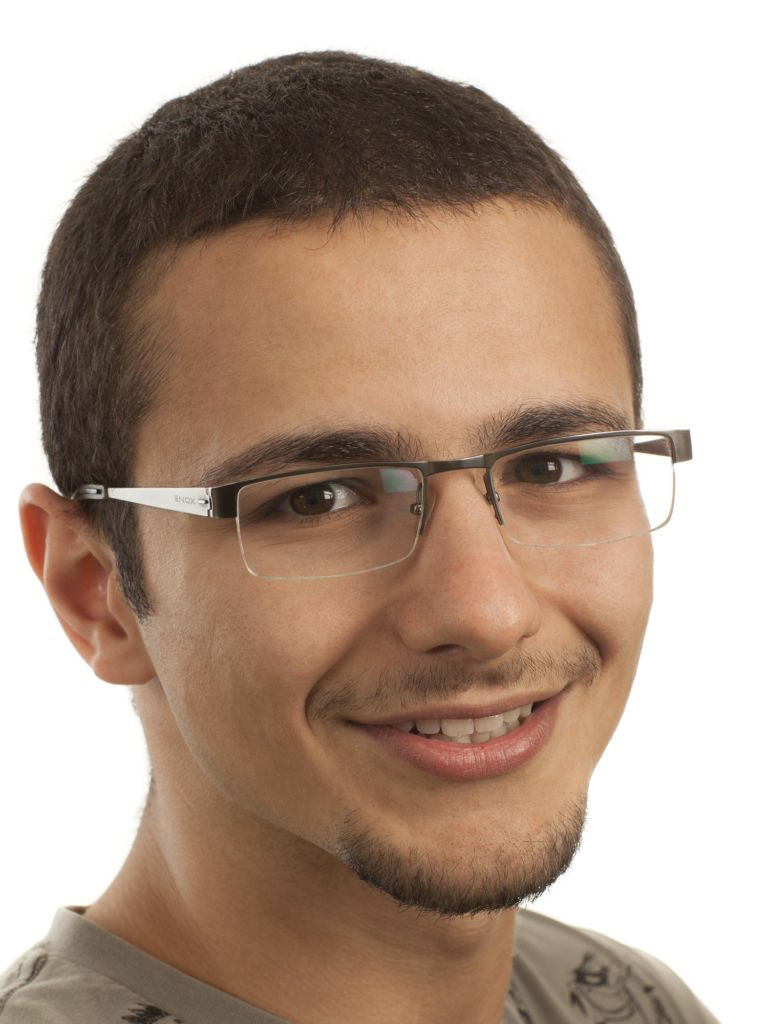}}]{Yaman Umuroglu}
received the
PhD degree from the Norwegian University of Science and Technology (NTNU), Norway and a joint
European MSc on Embedded Systems from the
Erasmus Mundus EMECS programme. He is a
research scientist at Xilinx Research Labs, Ireland.
His research takes a full-stack view of machine
learning with neural networks with a focus on highefficiency and high-performance implementations
and spans hardware-network codesign, techniques
for efficient arithmetic, sparsity, and quantization.
\end{IEEEbiography}

\begin{IEEEbiography}[{\includegraphics[width=1in,height=1.25in,clip,keepaspectratio]{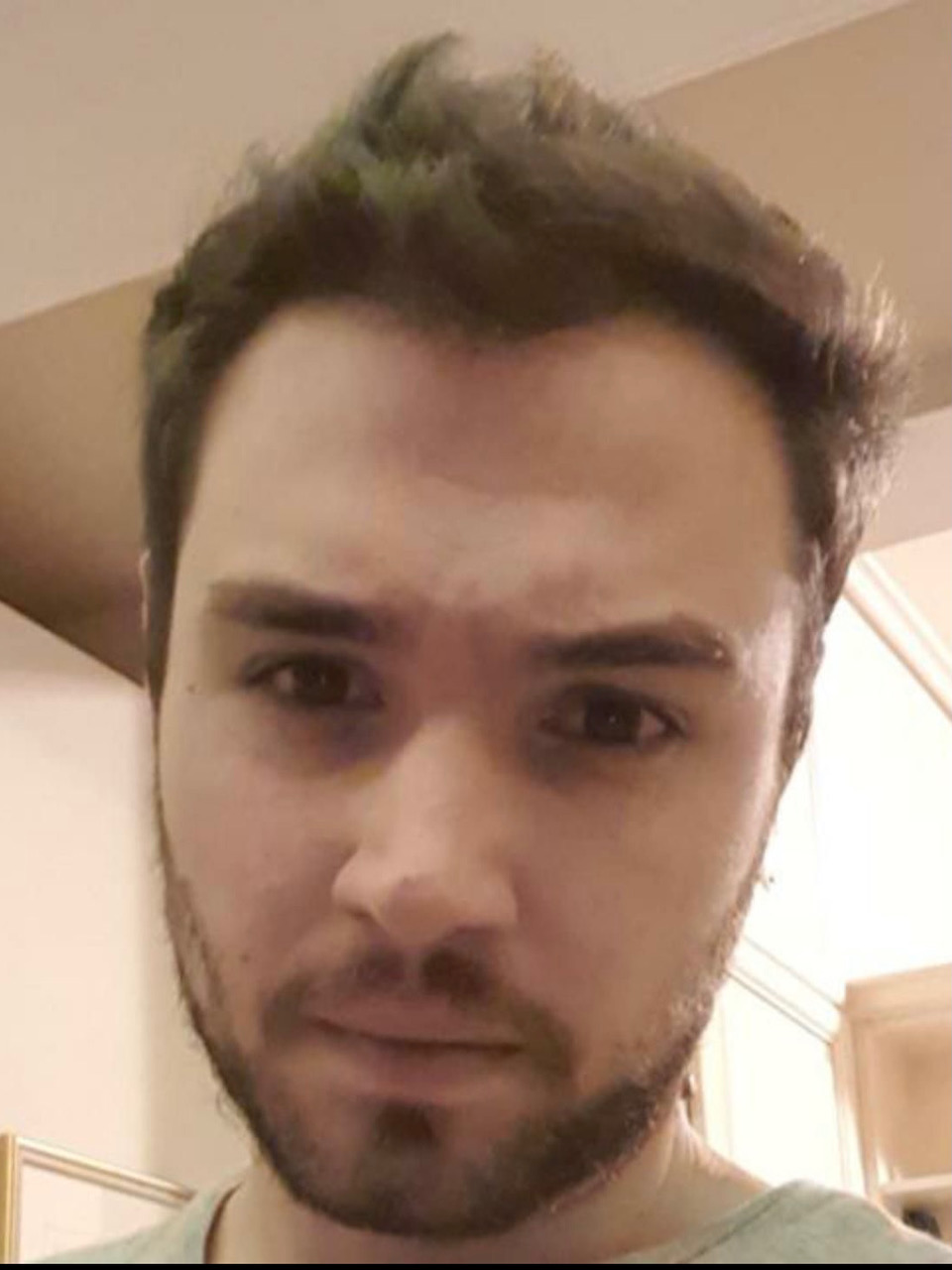}}]{Konstantinos Kanellopoulos}
is currently pursuing his PhD at ETH Zürich in the SAFARI research group. He completed his MEng and BSc at NTUA. His research interests lie at the intersection of software and hardware.
\end{IEEEbiography}

\begin{IEEEbiography}[{\includegraphics[width=1in,height=1.25in,clip,keepaspectratio]{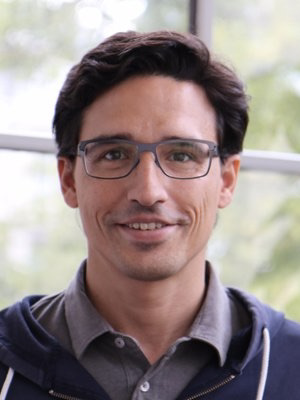}}]{Juan\ G\'omez-Luna}
is a senior researcher and lecturer at SAFARI Research Group @ ETH Zürich. He received the BS and MS degrees in Telecommunication Engineering from the University of Sevilla, Spain, in 2001, and the PhD degree in Computer Science from the University of Córdoba, Spain, in 2012. Between 2005 and 2017, he was a faculty member of the University of Córdoba. His research interests focus on processing-in-memory, memory systems, heterogeneous computing, and hardware and software acceleration of medical imaging and bioinformatics. He is the lead author of PrIM (\url{https://github.com/CMU-SAFARI/prim-benchmarks}), the first publicly-available benchmark suite for a real-world processing-in-memory architecture, and Chai (\url{https://github.com/chai-benchmarks/chai}), a benchmark suite for heterogeneous systems with CPU/GPU/FPGA.
\end{IEEEbiography}

\begin{IEEEbiography}[{\includegraphics[width=1in,height=1.25in,clip,keepaspectratio]{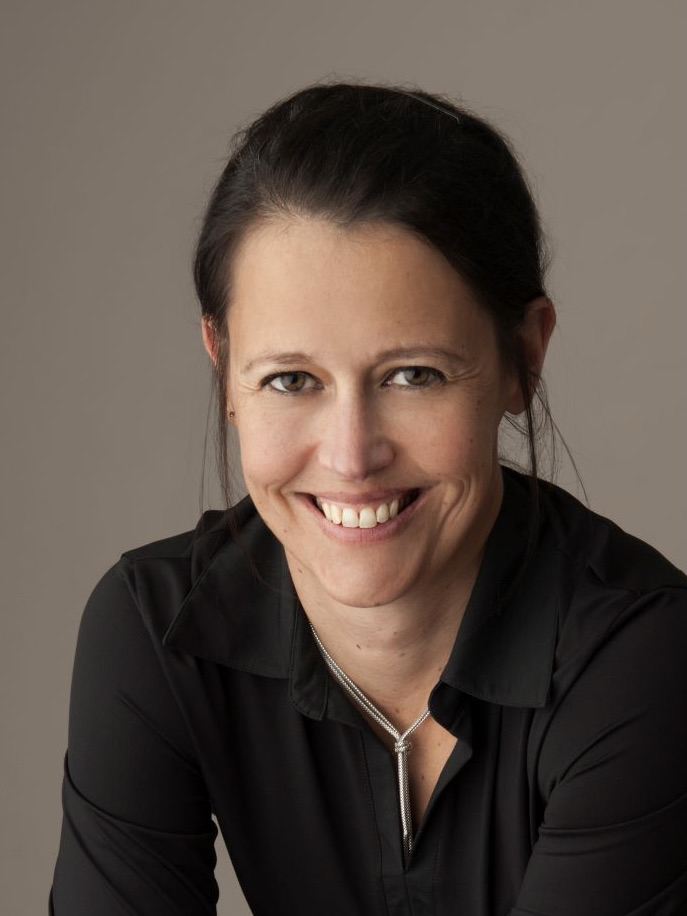}}]{Michaela Blott}
received the master’s degree from
the University of Kaiserslautern in Germany and
brings more than 25 years of computer architecture, FPGA and board design, in research institutions (ETH Zurich and Bell Labs) and development
organizations. She is a distinguished engineer at
Xilinx Research, Dublin, Ireland, where she heads
a team of international scientists driving exciting
research to define new application domains for
Xilinx devices, such as machine learning. She is
heavily involved with the international research
community serving as the technical co-chair of FPL’2018, workshop organizer (H2RC, ITEM’2020), and member of numerous technical program
committees (FPL, ISFPGA, DATE, etc.).
\end{IEEEbiography}

\begin{IEEEbiography}[{\includegraphics[width=1in,height=1.25in,clip,keepaspectratio]{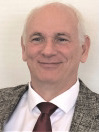}}]{Kees Vissers}
graduated from Delft University in the Netherlands. He worked at Philips Research in Eindhoven, The Netherlands, for many years. The work included Digital Video system design, HW–SW co-design, VLIW processor design and dedicated video processors. He was a visiting industrial fellow at Carnegie Mellon University, where he worked on early High Level Synthesis tools. He was a visiting industrial fellow at UC Berkeley where he worked on several models of computation and dataflow computing. He was a director of architecture at Trimedia, and CTO at Chameleon Systems. For more than a decade he is heading a team of international researchers at Xilinx in the CTO office. The research topics include machine learning applications and architectures, wireless applications, image processing applications and new datacenter applications. These applications drive next generation programming environments and architectures. He is a Fellow at Xilinx.
\end{IEEEbiography}

\begin{IEEEbiography}[{\includegraphics[width=1in,height=1.25in,clip,keepaspectratio]{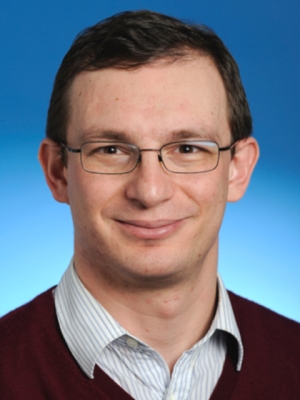}}]{Onur Mutlu}
 is a Professor of Computer Science at ETH Zürich. He is
also a faculty member at Carnegie Mellon University, where he
previously held the Strecker Early Career Professorship.  His current
broader research interests are in computer architecture, systems,
hardware security, and bioinformatics. A variety of techniques he,
along with his group and collaborators, has invented over the years
have influenced industry and have been employed in commercial
microprocessors and memory/storage systems. He obtained his PhD and MS
in ECE from the University of Texas at Austin and BS degrees in
Computer Engineering and Psychology from the University of Michigan,
Ann Arbor. He started the Computer Architecture Group at Microsoft
Research (2006-2009), and held various product and research positions
at Intel Corporation, Advanced Micro Devices, VMware, and Google.  He
received the IEEE High Performance Computer Architecture Test of Time
Award, the IEEE Computer Society Edward J. McCluskey Technical
Achievement Award, ACM SIGARCH Maurice Wilkes Award, the inaugural
IEEE Computer Society Young Computer Architect Award, the inaugural
Intel Early Career Faculty Award, US National Science Foundation
CAREER Award, Carnegie Mellon University Ladd Research Award, faculty
partnership awards from various companies, and a healthy number of
best paper or "Top Pick" paper recognitions at various computer
systems, architecture, and security venues. He is an ACM Fellow "for
contributions to computer architecture research, especially in memory
systems", IEEE Fellow for "contributions to computer architecture
research and practice", and an elected member of the Academy of Europe
(Academia Europaea). His computer architecture and digital logic
design course lectures and materials are freely available on YouTube
(\url{https://www.youtube.com/OnurMutluLectures}), and his research group
makes a wide variety of software and hardware artifacts freely
available online (\url{https://safari.ethz.ch/}). For more information,
please see his webpage at \url{https://people.inf.ethz.ch/omutlu/}.
\end{IEEEbiography}

\end{document}